\documentclass[10pt,twocolumn,letterpaper]{article}

\usepackage{cvpr}
\usepackage{times}
\usepackage{epsfig}
\usepackage{graphicx, subcaption}
\usepackage{amsmath}
\usepackage{amssymb}
\usepackage{xcolor}
\usepackage{booktabs}
\usepackage{makecell}
\usepackage{enumitem}
\usepackage[accsupp]{axessibility}
\usepackage{afterpage}
\usepackage{multirow}
\usepackage{hyperref}

\begin{document}

\title{ Underwater SONAR Image Classification and Analysis using LIME-based Explainable Artificial Intelligence }


     
\vspace{-1cm}

\author{Purushothaman Natarajan\\
Department of Computational Intelligence,\\
SRM Institute of Science and Technology\\
Kattankulathur, Tamil Nadu, 603203, India\\
{\tt\small \ c30945@srmist.edu.in}
\and
 Athira Nambiar\\
Department of Computational Intelligence,\\
Faculty of Engineering and Technology,\\
SRM Institute of Science and Technology\\
Kattankulathur, Tamil Nadu, 603203, India\\
{\tt\small \ athiram@srmist.edu.in}
}
\vspace{-1cm}
\maketitle
\vspace{-1cm}
\thispagestyle{empty}

\begin{abstract}
\vspace{-1cm}
Deep learning techniques have revolutionized image classification by mimicking human cognition and automating complex decision-making processes. However, the deployment of AI systems in the wild, especially in high-security domains such as defence, is curbed by the lack of explainability of the model. To this end, eXplainable AI (XAI) is an emerging area of research that is intended to explore the unexplained hidden ‘black box’ nature of deep neural networks.  This paper explores the application of the eXplainable Artificial Intelligence (XAI) tool to interpret the underwater image classification results, one of the first works in the domain to the best of our knowledge.  Our study delves into the realm of SONAR image classification using a custom dataset derived from diverse sources, including the Seabed Objects KLSG dataset, the camera SONAR dataset, the mine SONAR images dataset, and the SCTD dataset. An extensive analysis of transfer learning techniques for image classification using benchmark Convolutional Neural Network (CNN) architectures such as VGG16, VGG19, ResNet50, InceptionV3, DenseNet121, MobileNetV2, Xception, InceptionResNetV2, DenseNet201, NASNetLarge, NASNetMobile is carried out. On top of this classification model, a post-hoc XAI technique, \textit{viz.} Local Interpretable Model-Agnostic Explanations (LIME) are incorporated to provide transparent justifications for the model's decisions by perturbing input data locally to see how predictions change. Furthermore, Submodular Picks LIME (SP-LIME) a version of LIME particular to images, that perturbs the image based on the submodular picks is also extensively studied. To this end, two submodular optimization algorithms i.e. Quickshift and Simple Linear Iterative Clustering (SLIC) are leveraged towards submodular picks. The extensive analysis of XAI techniques highlights interpretability of the results in a more human-compliant way, thus boosting our confidence and reliability in the successful implementation of AI systems into the wild.
\vspace{-1cm}
\end{abstract}


\section{Introduction}
\label{introduction}
Imaging SONAR produces a reflectivity estimate of a portion of the ocean bottom using sound waves, hence, it is widely used to communicate, navigate, measure distances, and find objects on or beneath the water's surface~\cite{hansen2013synthetic}. In contrast to optical imaging, SONAR is preferred for underwater imagery because optical imaging systems rely on light conditions for imaging, but SONAR does not rely on ambient light. Therefore, it can operate effectively in low-light or even no-light conditions. Furthermore, SONAR can be used to identify the differences between the highlight and the shadow regions, which makes SONAR the best option for underwater surveillance and investigation whenever optical systems fail to do the job~\cite{fitzpatrick2020airborne}. Developments in autonomous underwater vehicles (AUVs) and remotely operated vehicles (ROVs) have enabled the implementation of underwater acoustic imaging equipment within a versatile framework on which the SONAR imaging equipment is placed below the underwater vehicles. 

Although AUVs can effectively collect data, they lack the capability to autonomously detect or categorize objects. Hence, it requires human assistance to manually review and categorize. For a naval base tasked with monitoring coastlines for potential threats such as enemy submarines, maritime monitoring has to be as quick as possible, but this multistage manual operation extends the response time and increases the threat of a foreign intruder and national security. In this regard, the automation of underwater image analysis on acoustic images can enhance the mission's autonomy while also reducing time, cost, and damage.

Traditional machine learning algorithms, such as Support Vector Machines (SVM)~\cite{hearst1998support} and decision trees~\cite{quinlan1987simplifying}, are employed for object recognition in SONAR images. However, the effectiveness of these shallow techniques is limited, yielding an accuracy of up to 92\%~\cite{xu2020svm},~\cite{febriawan2019support}. Later, deep learning algorithms like Convolutional Neural Networks (CNN) improved the accuracy by utilizing more computational resources, getting it to about 94.40\%~\cite{luo2019sediment}. The utilisation of transfer learning, wherein a pre-trained model serves as the base model, further boosted the accuracy to 97.33\%~\cite{chungath2023transfer}. 

Nevertheless, to apply such Artificial Intelligence (AI) systems in real-world scenarios of high-security concerns i.e. military/ defence, certain limitations curb the application of deep learning (DL) systems in the wild. One big challenge is the lack of availability of big data. Additionally, to come up with promising practical
solutions for such confidential data, it is quite essential to have the validation of the results by
conclusive interpretable evidence. However, this is where most of the existing DL models fail since
they act as a ‘black-box’ while predicting the results.
In such circumstances, it is crucial to enhance the reliability, trustworthiness, and interpretability of any deep neural network (DNN) model so that humans based on their situational awareness, can make decisions that involve interpreting the predictions from the model~\cite{gunning2021darpa}. This can help in detecting abnormalities, aiding in better decision-making, reducing false alarms, facilitating training, and maintaining transparency~\cite{broniatowski2021psychological},~\cite{neerincx2018using}. To this end, Explainable Artificial Intelligence (XAI) is an emerging area of research in Machine learning that is intended towards exploring the unexplained hidden “black box” nature of deep neural networks~\cite{gohel2021explainable}.

In this study, we address the black-box nature of deep learning model by incorporating XAI tools for underwater sonar image classification.  The proposed underwater XAI image classification pipeline consists of two stages: First, the performance of various backbone classification models such as VGG16, VGG19 ~\cite{simonyan2014very}, ResNet50 ~\cite{he2016deep}, Inception V3 ~\cite{szegedy2016rethinking}, DenseNet121, DenseNet201 ~\cite{huang2017densely}, MobileNetV2 ~\cite{sandler2018mobilenetv2}, Xception ~\cite{chollet2017xception}, InceptionResNetV2~\cite{szegedy2017inception}, NASNetLarge, and NASNetMobile~\cite{zoph2018learning} are analysed via Transfer learning technique. The significance of sampling techniques (random vs. stratified) as well as  data-balancing strategies (oversampling vs. undersampling) are also studied to see their impact on model performance.

Second, two popular explainable AI tools i.e. Local Interpretable Model-Agnostic Explanation (LIME) and submodular Picks LIME (SP-LIME) are employed to interpret the model decisions in a human-complaint manner. LIME and SP-LIME are perturbation-based techniques, suitable for most data types and images, respectively~\cite{ribeiro2016should}. Two different sub-modular optimization algorithms \textit{viz.} quickshift~\cite{vedaldi2008quick} and Simple Linear Iterative Clustering (SLIC)~\cite{achanta2012slic} are leveraged towards perturbing super-pixels in SP-LIME. Extensive analysis of the aforementioned XAI models are carried out to comprehend the model predictions, traits of the explainer predictions, impact of the hyperparameters on the explainer models and state-of-the-art comparisons. To the best of our knowledge, the literature on XAI for SONAR imagery analysis systems is scarce, making our model one of its early works in this field.

All the studies are carried out on a novel custom-made SONAR dataset curated in-house by consolidating various publicly available sonar datasets i.e. \textit{Seabed Objects KLSG} ~\cite{huo2020underwater}, \textit{SONAR Common Target Detection Dataset (SCTD)}~\cite{zhang2021self}, \textit{Camera SONAR dataset}~\cite{Terayama2018}, and \textit{Mine SONAR images}~\cite{mine_SONAR_images_dataset}. The key contributions of this paper are summarised as follows:
 
\begin{itemize}
    \item Development of a new tailored SONAR dataset by combining multiple publicly available SONAR image datasets.
    \item A comparative study to assess how well various benchmark models perform with Transfer Learning for SONAR image classification.
    \item  Brief study on the significance of data sampling techniques and data-balancing schemes during model training.
    \item Incorporating Explainable Artificial Intelligence (XAI) using LIME and SP-LIME, powerful tools offering visual explanations for underwater SONAR image classification results, one of its first kind to the best of our knowledge.
    \item Extensive quantitative and qualitative analysis on the explanation from various models,  the impact of sampling/data balancing strategies and hyperparameters and state-of-the-art comparison.
\end{itemize}

\noindent The forthcoming sections of this paper are organized as follows: The discussion of SONAR imaging and related work is presented in Section~\ref{background and related work}. The development of the custom dataset used for training is described in Section~\ref{dataset for learning}. Classification using pre-trained models and different Explainable Artificial Intelligence (XAI) methodologies, with a special focus on LIME and SP-LIME is described in Sections~\ref{methodology: transfer learning} and~\ref{experimental setup}, respectively. The experimental findings are showcased in Section~\ref{experimental results}. Finally, discussion and conclusion are drawn in Section~\ref{conclusion}.

\section{Background and Related work}
\label{background and related work}
In this section, we briefly describe the working principle of SONAR imaging, the literature review, and the motivation behind this research work.

\subsection{\textbf{SONAR working Principle}}
\label{sonar working principle}
Sound navigation and ranging (SONAR) uses sound waves to capture the scene on the ocean floor. SONAR has two major components, the transmitter and receiver~\cite{sutton1979underwater}. The transmitter transmits acoustic waves at a grazing angle to the bottom. These acoustic waves propagate through the water, bounce back by touching the objects or seafloor in their path, and then the receiver collects them, which illuminates a region on the seafloor. On the other hand, if there are no objects, then the back-scattered waves reach the transmitter, forming a uniform background. However, if objects are present on the seafloor, sound waves intercept and reflect from the objects and form a bright spot (commonly referred to as the ``highlight'') at a distance equal to the range of the object~\cite{elfes1987sonar}. By processing the signals, the SONAR system can generate detailed images, revealing the presence of underwater features such as reefs, wrecks, geological formations, and even enemy submarines, etc~\cite{blondel1997handbook},~\cite{hayes2009synthetic}.

\subsection{\textbf{Underwater Object Classification}}
\label{underwater object classification}
Different natural and artificial sounds, such as thermal noise, shipping noise, earthquakes, volcanic eruptions, and submarines, consistently disrupt the underwater acoustic environment. As a result, noise distorts and blurs the captured acoustic image~\cite{lurton2002introduction}. Furthermore, acoustic images generated by SONAR have a lower resolution compared to optical images, and this resolution varies depending on the distance between the sensor and the detected object~\cite{nguyen2019study}. Despite these challenges, the progress made in the field of automatic object detection in SONAR imagery has been impressive, considering the relatively short duration of time. Earlier versions of underwater object detection and classification systems depended on a two-step detection process~\cite{daniel1998side}. The first step involves collecting data, while in the second step, humans manually classify the collected data using predefined classes based on the available templates~\cite{myers2010template}.

Langner et al.~\cite{langner2009side} utilised probabilistic neural networks and pattern-matching techniques based on clustering algorithms for automatic object detection and classification on side scan sonar images. Myers et al.~\cite{myers2010template} used template matching techniques to compare the target signature made from a simple acoustic model with a picture of the object that needs to be classified. H.Xu et al.~\cite{xu2020svm} developed an Adaboost cascade classification model based on support vector machines (SVM) and achieved an accuracy of about 92\%. The statistical approach~\cite{abu2019statistically} utilises the HS-LRT (highlight-shadow likelihood ratio test) to identify and emphasise areas with both shadow and highlight and subsequently, the region of interest is determined based on these shadow and highlight regions. Finally, an SVM is employed to detect the desired regions of interest.

In 2017, Mckay et al.~\cite{8232162} began utilising deep neural networks for SONAR image classification and also utilised transfer learning with the pre-trained model VGG16 and got an impressive accuracy of 93\% on their custom side scan SONAR (SSS) dataset, and improved the accuracy further to 97.7\% by including semisynthetic data generated from optical images. In 2018, Fuchs et al.\cite{fuchs2018object} also utilized transfer learning for sonar image classification on aracati dataset with ResNet-50 as a backbone model and also did a comparative study with traditional machine learning approaches and achieved an accuracy of 88\%.  In 2020, Pannu et al.~\cite{pannu2020deep} utilised CNN for the classification of synthetic aperture radar (SAR) images, adeptly addressing data limitations through strategic augmentation. Transitioning to the M-STAR dataset, a deep learning-based approach achieves an impressive 80\% accuracy, further boosted to 97\% after meticulous balancing.

The comparative study by Du et al.~\cite{du2023comparative} leveraged transfer learning for sonar image classification on the Seabed Objects KLSG dataset with the pre-trained models, namely AlexNet, VGG-16, GoogleNet, and ResNet101, which achieved a maximum accuracy of 94.81\%. A study by Chungath et al.~\cite{chungath2023transfer} also used transfer learning with a pre-trained model ResNet-50 to classify sonar images, achieved an impressive accuracy of 97.33\%, and also showed the ways to train an image classification model with very few images using few-shot learning-based deep neural networks.

The comprehensive survey by Steiniger et al.~\cite{steiniger2022survey}, Domingos et al.~\cite{domingos2022survey}, and Neupane et al.~\cite{neupane2020review} explored past and current research involving deep learning for feature extraction, classification, detection, and segmentation of side scan and synthetic aperture SONAR imagery. Their work provides an insightful overview, serving as a cornerstone for understanding the dynamic applications of computer vision in this specialised underwater water domain.

\subsection{\textbf{Explainable AI for SONAR Image Analysis}}
\label{xai for sonar image analysis}
Most of the existing deep learning models are black boxes in nature whose predictions cannot be interpreted or understood by humans. The integration of explainable AI offers a transparent lens, providing nuanced visual explanations for the predictions~\cite{gohel2021explainable}. It helps verify the performance of the deep learning models and also enhances user trust in those models. Nguyen et al. conducted a thorough comparison study that looked at how well explainable AI approaches such as LIME~\cite{ribeiro2016should}, SHAP~\cite{lundberg2017unified}, and Grad-CAM~\cite{selvaraju2017grad} work in a wide range of industrial settings with images~\cite{nguyen2021evaluation}.
Further, in other domains such as remote sensing, medicine, etc., XAI techniques are employed to verify the performance of deep learning models~\cite{bhandari2022explanatory},~\cite{kakogeorgiou2021evaluating}. Hassan et al.~\cite{hassan2022prostate} used XAI on ultrasound and MRI images to classify prostate cancer in men. It is also used to study several respiratory disorders, including COVID-19, pneumonia, tuberculosis and skin lesion classification, to help doctors and clinicians come to strong and logical conclusions about deep learning models for detection and classification~\cite{bhandari2022explanatory},~\cite{nigar2022deep}. Bennet et al.~\cite{bennetot2022greybox} created a neural-symbolic learning framework utilizing a CNN model with a latent space predictor to classify images. This framework gives information about the predicted image through an explainable latent space. Similarly, it is also used to explain the predictions of complex image captioning models, which visually depict the part of the image corresponding to a particular word in the caption through a visual mask~\cite{sahay2021approach}.

The XAI programme ~\cite{gunning2019darpa} by DARPA underscores the ongoing efforts to enhance understanding, trust, and management of AI systems in the realm of defence. During the period of this programme, from 2015 to 2021, 11 XAI teams explored various machine learning approaches, i.e. tractable probabilistic models, causal models and explanation techniques leveraging state machines generated by reinforcement learning algorithms, Bayesian teaching, visual saliency maps and GAN dissection. ~\cite{gunning2021darpa}. Recent work on deep learning-based explainable target classification was studied by Pannu et al.~\cite{pannu2020deep}. In that work, the explainable AI tool LIME was leveraged to verify the performance of the synthetic aperture radar(SAR) image classification models.

There are very few research works with explainable AI for underwater SONAR imagery. To the best of our knowledge, the only known work in the domain is by~\cite{walker2021explainable}, wherein Walker et al. developed a LIME-based XAI classification model for seafloor texture classification. Contrary to that work, our work addresses a reliable multi-class classification model by comparing different backbone models using transfer learning approach, and also by addressing data balancing and sampling strategies. Additionally, XAL techniques LIME and SP-LIME are incorporated to interpret the predictions of the top-performing model, in a human-compliant manner. Further, detailed quantitative and qualitative analyses on the impact of different hyper-parameters and computational constraints are also investigated in this work.

\section {Dataset for Learning}
\label{dataset for learning}
The foundation of any machine learning model is the training dataset. The publicly available SONAR dataset is limited mainly because SONAR, primarily utilized for defence applications, generates sensitive images that are kept confidential. Hence, initial studies in SONAR image classification employed private datasets of underwater acoustic images. Therefore, the gathering of data is vital in this scenario. This section details the experimental setup for acquiring and preparing the dataset for our studies.

\begin{table*}[t!]
    \begin{center}
    \footnotesize
    \captionsetup{font=footnotesize}
    \caption{Dataset samples: First row: class name; second row: number of images per class; the remaining corresponding rows are the sample images from that particular class.}
    \begin{tabular}{l}
    \includegraphics[width=0.95\textwidth]{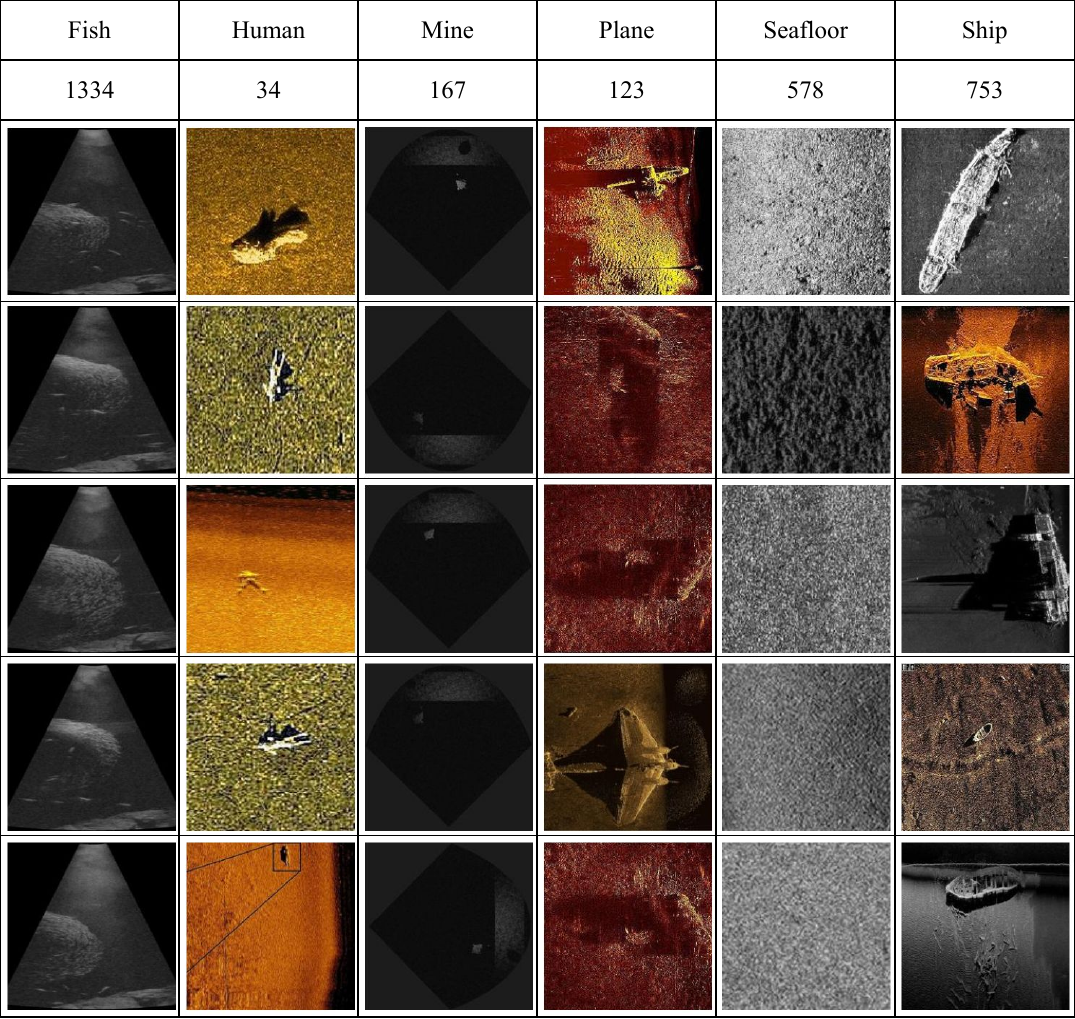} \\
    \end{tabular}
    \label{tbl:dataset preview}
    \end{center}
    \vspace{-0.75cm}
\end{table*}

\subsection{Data Acquisition}
\label{data acquisition}
Acquiring data in an underwater environment necessitates coordinated efforts from multiple agencies and departments. The collection and labelling of a substantial amount of acoustic data for educational purposes is an expensive and time-consuming process. The confidential nature of the SONAR data results in sparse publicly available datasets with very few classes. Hence, we develop a new custom-made dataset by consolidating from four different publicly available open-source SONAR datasets, listed as follows:

\begin{enumerate}
    \item The primary dataset, referred to as \textbf{\textit{Seabed Objects KLSG}} ~\cite{huo2020underwater}, comprises 1190 Side Scan SONAR images with a total of 385 images of shipwrecks, 36 images of drowning victims, 62 images of planes, 129 images of mines, and 578 images of the seafloor. The authors gathered these real SONAR images with the assistance of various commercial SONAR suppliers, including Lcocean, Klein Martin, and EdgeTech accumulated over a period of ten years. A few classes with a total of 1171 images with a variety of dimensions, along with augmented images excluding mines, are accessible to the public through the link~\cite{seabedobjects}, with the purpose of supporting academic research in this particular domain.

    \item The second dataset named \textbf{\textit{Sonar Common Target Detection Dataset (SCTD)1.0}}, was developed by Tsinghua University, Fudan University, Shanghai University, Hohai University, Jiangxi University of Technology, et al.~\cite{zhang2021self}, consists of a total of 596 images belonging to 3 classes. However, only 317 images, specifically 57 images of planes, 266 images of shipwrecks, and 34 images of drowning victims, each with different dimensions, are publicly accessible through the link~\cite{sctd}.

    \item The \textbf{\textit{Camera SONAR dataset}}~\cite{Terayama2018}, published by Kei Terayama et al. consists of 1334 underwater SONAR and camera photos of fishes with a dimension of 256x512. For this study, we will only be using the SONAR images.

    \item The dataset referred to \textbf{\textit{Mine SONAR images}} provided in the Roboflow universe by Phan Quang Tuan comprises 167 SONAR images depicting mines with a dimension of 416x416~\cite{mine_SONAR_images_dataset}.
\end{enumerate}

By combining the aforementioned datasets, we curated our tailor-made dataset, which comprises a total of 2989 side scan SONAR images with 1334 fish images, 34 human images, 167 mine images, 123 plane images, 578 seafloor images, and 753 ship wreck images. Some sample images of our custom dataset, together with the image count for each label, are provided in Table~\ref{tbl:dataset preview}. All images in the dataset are obtained from the originally captured images without any preprocessing.

\subsection{Data Splitting}
\label{data splitting}

In a supervised learning environment, the objective is to construct a model that can accurately predict both the input sample and unseen samples. To evaluate the model's performance, we divide the dataset into three mutually exclusive sets: train, validation, and test sets~\cite{reitermanova2010data}. The model is developed using the training set, and then the developed models are fine-tuned using the validation set. Finally, the model's performance is evaluated using the test dataset. Splitting the dataset has a major impact on the model's quality by reducing both bias and variance~\cite{cody2022systematic}. A poor data split might lead to poor model performance, particularly in imbalanced datasets such as the one in the current case. This problem of data imbalance is common in many real-time applications where the number of data points in some classes is significantly higher than the other classes. 

\begin{figure}[h!]
    \centering
    \footnotesize
    \includegraphics[width =0.45\textwidth]{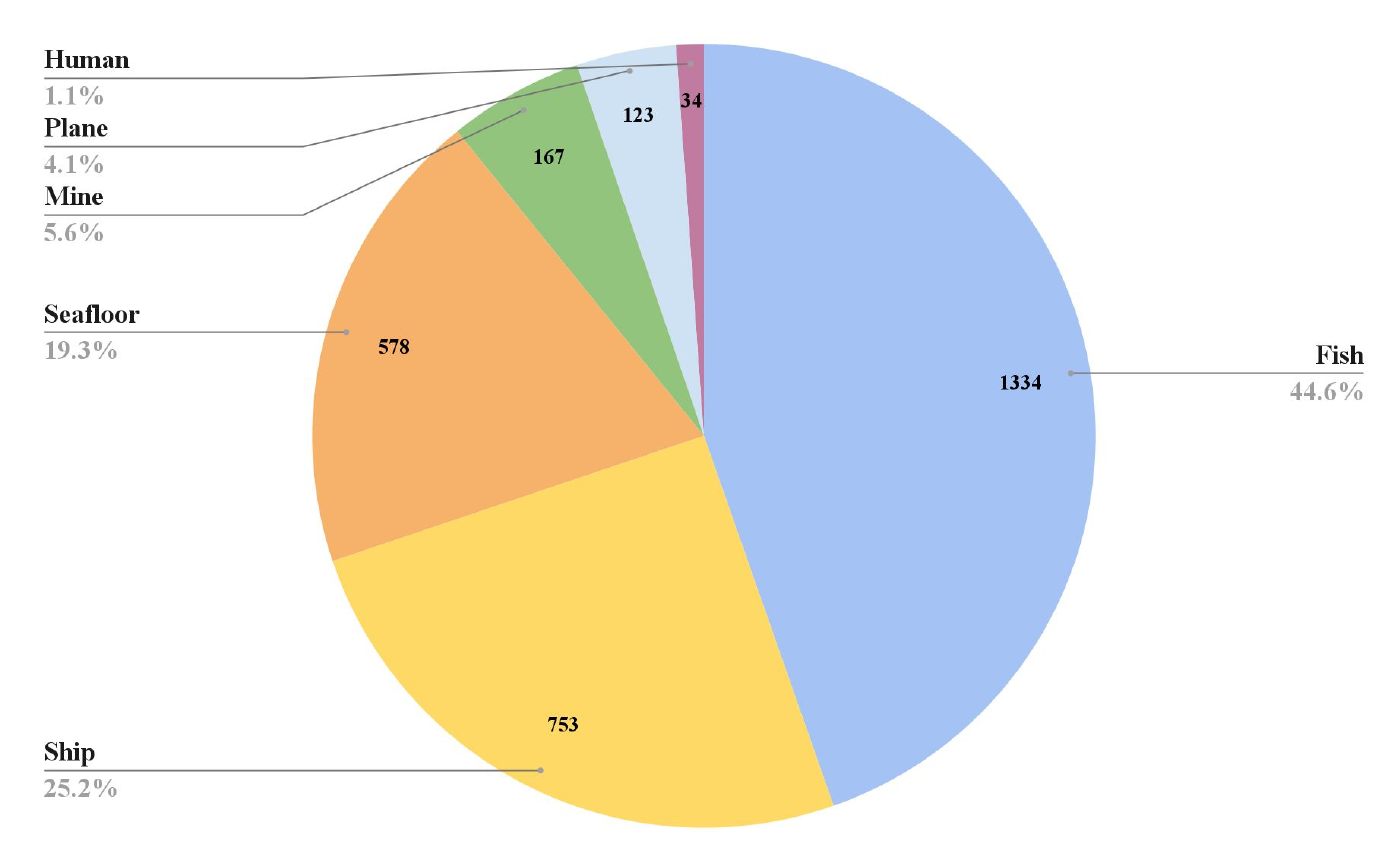}
    \begin{center}
    \captionsetup{font=footnotesize}
    \vspace{-0.25cm}
    \caption{Class-wise distribution of the custom dataset. The class Fish
contributes to around 45\% of the dataset.}
    \label{fig:class-wise distribution}
    \end{center}
    \vspace{-0.75cm}
\end{figure}

In our case, the majority class has 1334 images, and the minority class has only 34 images. In this situation, we should ensure that the train and test split have approximately equal numbers of samples in all the classes to maintain the class distribution intact. Otherwise, the model might overfit. Fig.~\ref{fig:class-wise distribution} represents the class-wise distribution of our custom dataset. As seen, the fish class dominated with 44.6\% of the dataset, while the human class represented only 1.1\% of the dataset.

The task of splitting the dataset for a deep learning network can be seen as a sampling problem, where the given dataset D containing N data samples needs to be partitioned into three mutually disjoint subsets, so we split the dataset to train, validation and test datasets. As per~\cite{reitermanova2010data}, if the dataset is imbalanced, simple random sampling cannot represent and maintain the class distribution for all the classes in the dataset, but statistical sampling-based data splitting approaches, such as stratified sampling~\cite{cochran1977sampling}, work better over random sampling because the core idea is to inspect the internal structure and distribution of each subclass of the dataset D and utilise this information to divide the dataset into a set of relatively homogeneous sample groups. In the current problem, both simple random sampling and stratified sampling are applied to the available dataset in the ratio of 70:15:15 among train, validation, and test splits, with simple random sampling randomly selecting images and stratified sampling selecting an image from a class with probability inversely proportional to the size of that class.

\section{Methodology: Transfer Learning-Based
Image Classification, Data Balancing and LIME-based Explanation}
\label{methodology: transfer learning}

Training a deep neural network (DNN) from scratch with random weights demands substantial computational resources, extensive datasets, and considerable processing time~\cite{weiss2016survey}. In the other hand, Transfer learning trains a deep neural network (DNN) with less data and in less time all while getting better results~\cite{ribani2019survey}. This is beneficial when training on intensive image datasets, where the challenges associated with collecting and labelling data, given its cost and time-intensive nature, and privacy concerns surrounding real user data make its utilization increasingly challenging~\cite{sarkar2018hands}.

\subsection{\textbf{Transfer Learning}}
\label{transfer learning}
Transfer learning facilitates the rapid prototyping of new machine learning models by leveraging the weights and biases of pre-trained models from a source task~\cite{deng2009large}. This transfer of weights, essentially the transfer of knowledge, offers notable benefits in several ways, not only in speeding up the convergence process during training but also in establishing a higher initial accuracy. This is especially advantageous when dealing with smaller datasets, demonstrating the efficiency and effectiveness of transfer learning~\cite{chungath2023transfer}.

Consider a domain, denoted as \(D\), with a feature space \(X\), and a marginal probability distribution \(P(X)\), where \(x\) is a set \(\{x_1, x_2, \ldots, x_n\}\) belonging to \(X\). Represented as \(D = \{X, P(X)\}\), let \(T\) be a task defined as \(\{Y, f(x)\}\), where \(Y\) is the label space and \(f(x)\) is the target predictive function.
Suppose \(Ds\) and \(Ts\) as the source domain and corresponding learning task, and \(Dt\) and \(Tt\) as the target domain and target learning task, respectively. Let \(Ds = \{(xs_1, ys_1), (xs_2, ys_2), \ldots, (xs_n, ys_n)\}\) encompass the elements of the source domain data, where each \(xs_i \in Xs\) represents a data point and \(ys_i \in Ys\) denotes its class label. We want to make the target predictive function \(f_T(.)\) better for the learning task \(Tt\) by using data from \(Ds\) and \(Ts\), where \(Ds \neq Dt\) or \(Ts \neq Tt\).

The transfer learning process begins with feature extraction from both the source domain \(Ds\) and the target domain \(Dt\). Instead of initializing the network with random weights and biases, the idea is to initialize the weights from the source domain (\(Ws\)) and adjust the weights to the target domain \(Wt\). The target weights \(Wt\) are optimized by backpropagation and gradient descent using the target data \(Dt\) and are utilized it to predict the target learning task (\(Tt\)). According to Chungath et al.~\cite{chungath2023transfer} and Ribani et al.~\cite{ribani2019survey}, this process involves making small changes to the network weights so that the model can understand high-level features that are unique to the dataset. This leads to better performance than direct training. 

\begin{figure*}[h!]
    \vspace{-0.5cm}
    \centering
    \footnotesize
    \includegraphics[width=\textwidth]{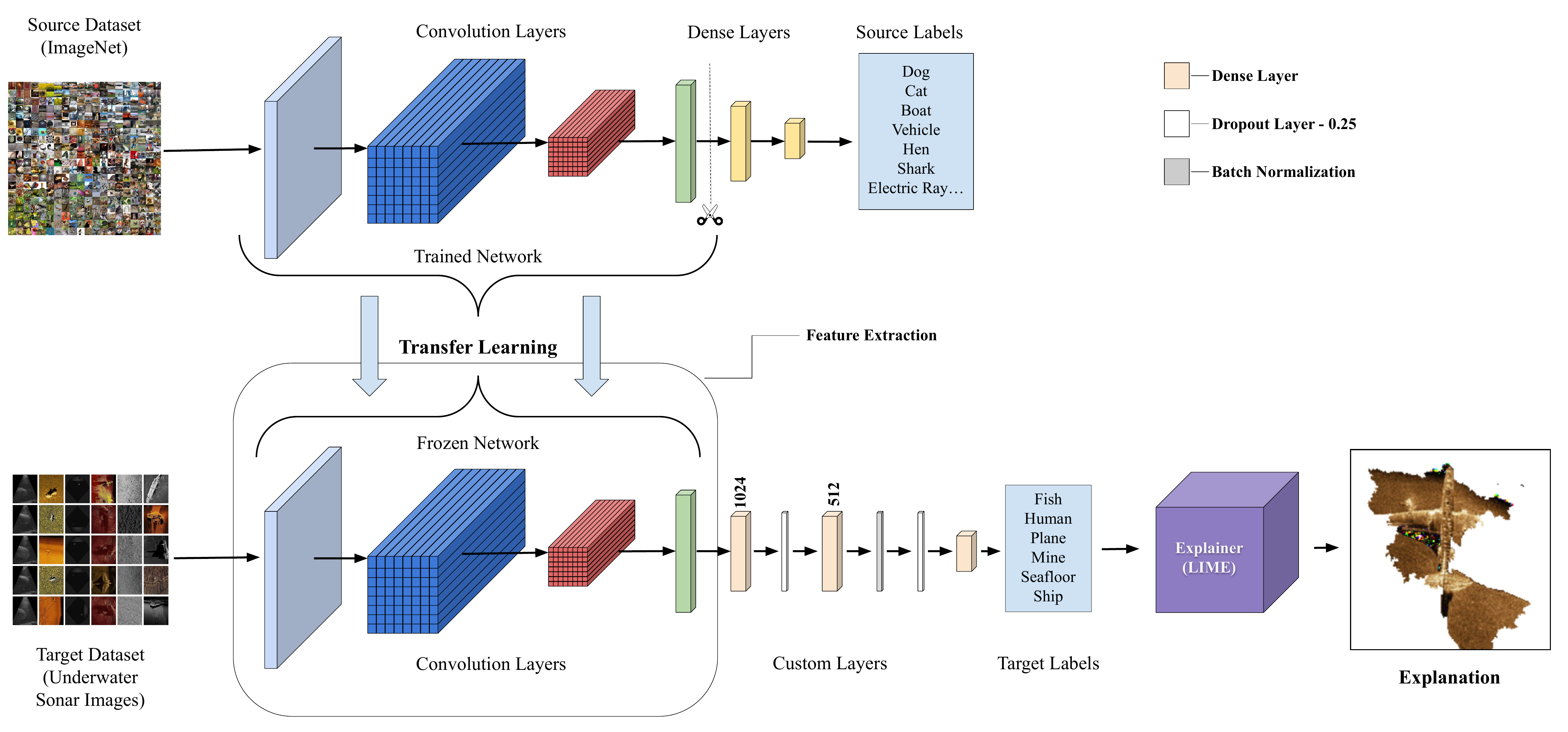}
    \centering
    \captionsetup{font=footnotesize}
    \caption{Overall architectural diagram of the proposed Underwater SONAR Image Classification model using Transfer learning and LIME-Explainable AI.}
    \label{fig:Transfer Learning for Image Classification with LIME}
    \vspace{-0.35cm}
\end{figure*}

Refer to the visual representation of Fig.~\ref{fig:Transfer Learning for Image Classification with LIME} for the overall architecture of the proposed underwater SONAR image classification model using
transfer learning and LIME. The model weights are pre-trained on ImageNet followed by fine-tuning on a target SONAR dataset. The target model is customized with three trainable layers, two dropouts, and a normalization layer. Additionally, an explainer (LIME/ SP-LIME) is incorporated on top of the classifier to provide explanations for the predictions made by the classification model.

\subsection{\textbf{Handling Data Imbalance}}
\label{handling data imbalance}

\begin{figure}[ht!]
    \centering
    \includegraphics[width=0.45\textwidth]{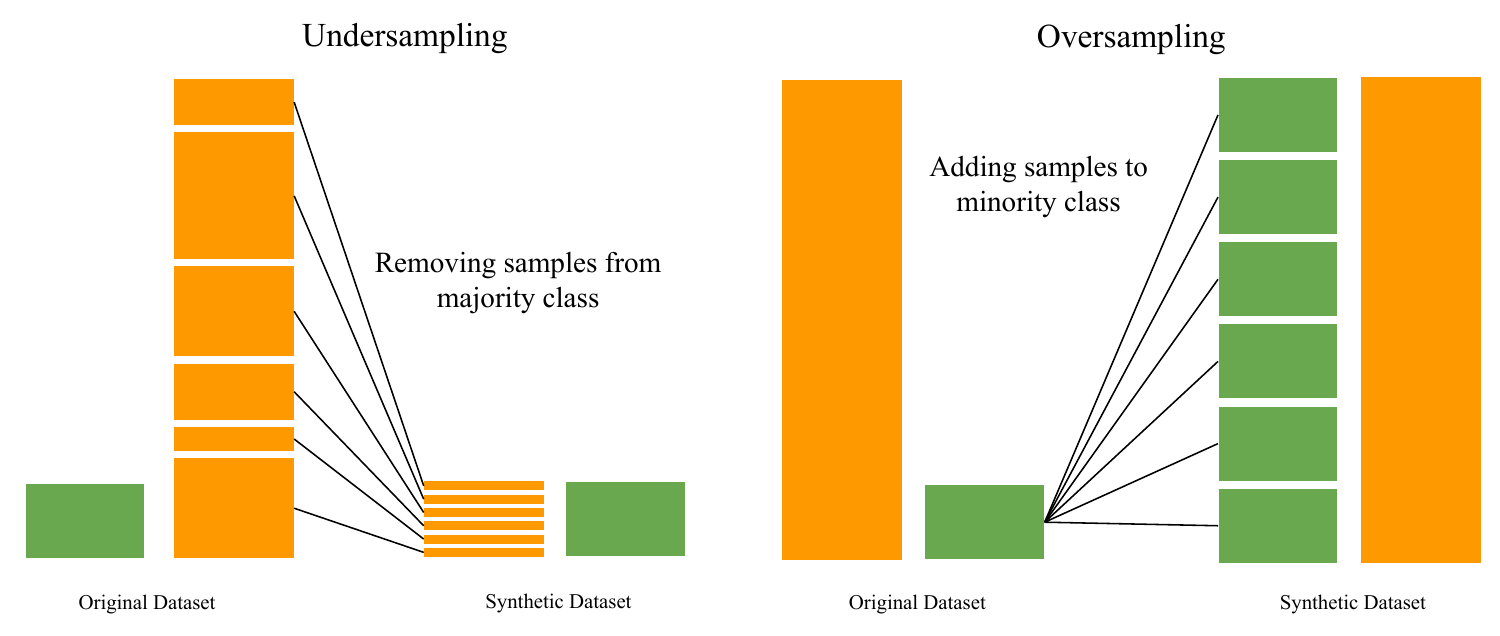}
    \captionsetup{font=footnotesize}
    \caption{Oversampling and Undersampling Illustration}
    \label{fig:Oversampling and Undersampling Illustration}
    \vspace{-0.35cm}
\end{figure}

\begin{figure*}[h!]
\vspace{-0.5cm}
    \centering
    \includegraphics[width=\textwidth]{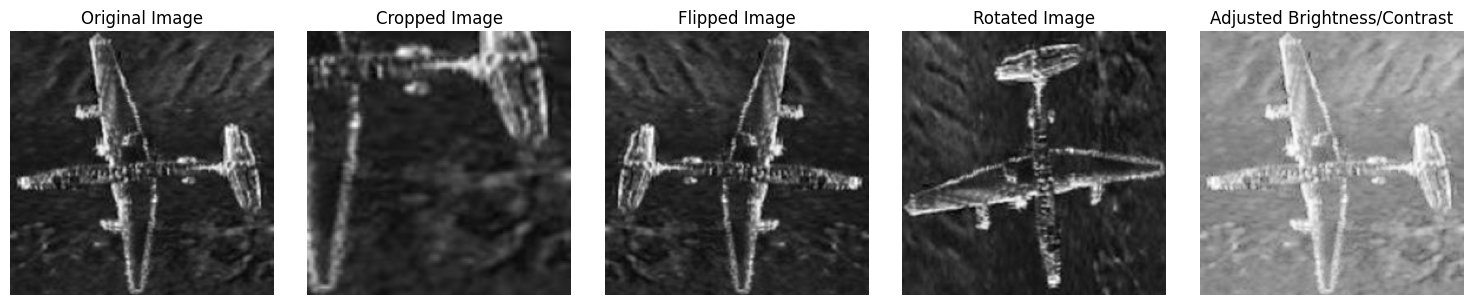}
    \begin{minipage}{\textwidth}
            \captionsetup{font=footnotesize}
        \caption{Augmented Images}
        \label{augmented image}
    \end{minipage}
    \vspace{-0.75cm}
\end{figure*}

As explained in Section~\ref{data splitting}, there exists a class imbalance problem in the dataset due to high variance in the number of samples per class. This issue is tackled through a dual strategy: Oversampling and Undersampling techniques ~\cite{kotsiantis2006handling} (refer Fig.\ref{fig:Oversampling and Undersampling Illustration}). Oversampling refers to the technique of adding more images to minority classes, whereas undersampling refers to the technique of randomly discarding some images from the majority class~\cite{mohammed2020machine}. In our case, the majority classes are fishes, seafloor, and ships, while the minority classes are plane, human, and mine. To address the scarcity of plane, human, and mine images. We employ image augmentation techniques such as flipping, rotation, cropping, adjusting the brightness, contrast, random erasing, noise injection, mixing images, adding blur, sharpness, and colour jittering to generate synthetic images, as an instance of oversampling~\cite{shorten2019survey} (Refer to Fig.\ref{augmented image}). Conversely, to harmonise the abundance of ship, fish, and seafloor images, we randomly discard some with equal probabilities to balance image numbers for each class as an undersampling approach~\cite{matsuoka2021classification}, ~\cite{kotsiantis2003mixture}. The number of images before and after augmentation is mentioned in Table~\ref{tbl:number of imageg per class}.

However, it's crucial to note that Kotsiantis et al.~\cite{kotsiantis2006handling} and Chawla et al.~\cite{chawla2002smote} cautioned against random oversampling due to its potential to heighten the risk of overfitting. This arises from the replication of exact copies of minority class examples, leading to the construction of seemingly accurate rules that may inadvertently be overly specific to individual replicated instances. Furthermore, in scenarios where the dataset is already sizeable but imbalanced, oversampling can introduce an additional computational burden. This underscores the importance of a nuanced approach to navigating the trade-offs associated with image oversampling techniques in the context of imbalanced datasets.

\vspace{0.4cm}
\begin{table}
    \centering
    \footnotesize
    \renewcommand{\arraystretch}{1.2}
    \captionsetup{font=footnotesize}
    \caption{Number of Images/Class: First column: number of images before augmentation; \& Second column: number of images after augmentation.}
    \begin{tabular}{|c|c|c|}
        \hline
        \textbf{Class} & \textbf{Before Augmentation} & \textbf{After Augmentation} \\
        \hline
        Fish & 1334 & 498 \\
        Human & 34 & 498 \\
        Mine & 167 & 498 \\
        Plane & 123 & 498 \\
        Seafloor & 578 & 498 \\
        Ship & 753 & 498 \\
        \hline
        \textbf{Total} & \textbf{2989} & \textbf{2988} \\
        \hline
    \end{tabular}
    \vspace{-0.5cm}
    \label{tbl:number of imageg per class}
\end{table}

\subsection{\textbf{eXplainable AI (XAI)}}
\label{section - xai}

As explained earlier, deep learning models are opaque by nature; they only predict the target with a probability score. To interpret such `black-box' models, Explainable AI (XAI) tools facilitates in gaining human reliance on AI technology. XAI addresses bias understanding and fairness by navigating the bias-variance trade-off in AI/ML models~\cite{arrieta2020explainable}. This promotes fairness and aids in mitigating bias during the justification or interpretation phase. Recognizing and managing bias is crucial for ethical AI practices, preventing the unjust distribution of benefits to any party. Thus, XAI enhances the transparency, fairness, and trustworthiness of AI systems, providing meaningful insights into the decision-making process for both experts and non-experts alike.

\noindent \textbf{XAI Techniques} can be broadly categorised into `transparent’ and `post-hoc' methods~\cite{gohel2021explainable}. Transparent approaches, like decision trees~\cite{mahbooba2021explainable} and random forest~\cite{neto2020explainable}, are naturally clear and easy to understand, which makes it easier to fully understand how the model works. These methods are well-suited for simpler models, offering direct insights into the decision-making process ~\cite{robnik2018perturbation}. 
On the other hand, post-hoc approaches explain the models' decision without altering or even knowing the inner works. Examples of such methods are Grad-CAM~\cite{zhou2016learning}, SHAP (Shapley Additive Explanations)~\cite{lundberg2017unified}, and LIME (Local Interpretable Model-Agnostic Explanations)~\cite{ribeiro2016should}. These techniques are particularly useful for understanding complex interactions in nonlinear or black-box models because they offer localised insights that improve transparency when dealing with complex data.

\noindent \textbf{Traits of an Explainer:} An explainer should be made considering the following characteristics: reliability, interpretability, and explainability based on psychological constructs as described in~\cite{broniatowski2021psychological}.
 \textbf{Reliability}, also called fidelity in simple terms, means the explanation must be a reasonable representation of what the system does internally~\cite{ribeiro2016should}.
\textbf{Interpretability} refers to the ability to understand the concepts learned by a model~\cite{narayanan2018humans}, 
\textbf{Explainability} refers to the ability to justify the decisions made by a model and accurately describe the mechanism, or implementation, that led to an algorithm’s output~\cite{broniatowski2021psychological}. Explainability is important for gaining the trust of users, providing insights into the causes of decisions, and auditing the model’s behaviour.
Interpretability and explainability are two related but different concepts in the context of machine learning models, which is always debatable, as mentioned in~\cite{phillips2020four}.

\subsection{\textbf{Local interpretable Model-Agnostic Explanations (LIME) as an XAI tool}}
\label{section: lime as a xai tool}

Local Interpretable Model-Agnostic Explanations (LIME) is a novel explanation technique used to explain the predictions made by any classifier in a manner that is both interpretable and faithful by training a locally interpretable model around the prediction~\cite{ribeiro2016should}. It provides faithful explanations for predictions made by any classifier or regressor. This is achieved through the local approximation of the model, enabling the use of an interpretable model for enhanced understanding and interpretability~\cite{ribeiro2016should}. In other words, LIME improves our model's interpretability by tweaking input data locally to see how predictions change. Tweaking the data locally explains a single prediction for that instance, while global tweaking looks at the model's behaviour across all data. The idea behind local explanations is to create an interpretable representation of an underlying model, making its decision-making process more understandable.

The mathematical formulation of the LIME explainer is detailed below:
Let \(f: \mathbb{R}^d \rightarrow \mathbb{R}\) denote the model to be explained. $x \in \mathbb{R}^d$ is the original representation of the instance being explained. In a classification context, \(f(x)\) represents the probability (or a binary indicator) that the instance \(x\) belongs to a specific class. Additionally, let \(\pi_x(z)\) be a proximity measure between an instance \(z\) and \(x\), defining the locality around \(x\). Consider the loss function \(L(f, g, \pi_x)\) as a representation of how accurate the explanation function \(g\) is at approximating \(f\) within the locality defined by \(\pi_x\). The loss function $L$ for locally weighted squared loss is expressed as in \textcolor{black}{(\ref{equ: loss fuction})}

\vspace{-0.25cm}
\begin{equation}
L(f, g, \pi_x) = \sum_{z, z_0 \in Z} \pi_x(z) \cdot (f(z) - g(z_0))^2 
\label{equ: loss fuction}  
\vspace{-0.15cm}
\end{equation}

where $Z$ is a dataset containing sampled instances $z$, and $f(z)$ represents the labels obtained from the original model. The term $z_0$ is a binary indicator of perturbed samples used to generate the original representation of instances. The function $g(z_0)$ is selected to be a class of linear models, specifically $g(z_0) = w_g \cdot z_0$. This formulation captures the essence of the locally weighted squared loss, emphasizing the importance of each sampled instance based on its proximity to the original instance $x$ in the feature space. The term $\pi_x$ in \textcolor{black}{(\ref{equ: loss fuction})} functions as a locality measure and is computed as follows\textcolor{blue}{,}

\begin{equation}
\vspace{-0.15cm}
\pi_x(z) = \exp \left( - \frac{D(x, z)^2}{\sigma^2} \right) 
\label{equ: pi_z}
\end{equation}

where $D$ serves as the distance metric (e.g., cosine, euclidean distance) between the two instances $x$ and $z$. The bandwidth of the kernel $\sigma$ is a parameter that requires careful selection. By resolving the following optimization \textcolor{black}{(\ref{equ:distance metric})}, LIME determines the explanation it produces, where $\xi$ represents the local behavior of the complex model $f$ around the instance $x$.

\begin{equation}
\xi_(x) = \arg\min_{g \in \mathcal{G}} L(f, g, \pi_x) + \Omega(g) 
\label{equ:distance metric}
\end{equation}

where $L(f, g, \pi_x)$ represents the loss function measuring how well interpretable model $g$ approximates black-box model $f$ for a perturbed sample $\pi_x$ and $\Omega(g)$ is the complexity measure quantifying the simplicity of interpretable model.

To ensure both interpretability and reliability, the goal is to minimize the loss function \(L(f, g, \pi_x)\), while restricting the complexity of the explanation \(\Omega(g)\) to a level that is interpretable by humans. This is achieved by penalizing the explanation function \(g\) if it becomes excessively complex in order to maintain the complexity of the surrogate model within acceptable limits.

\subsection{\textbf{Submodular Picks Local interpretable Model-Agnostic Explanations (SP-LIME) for images}}
\label{sp-lime for images}

The classifier may represent the image as a tensor with three colour channels per pixel, but an interpretable representation for our image classification task might be a binary vector indicating the ``presence" or ``absence" of a contiguous patch of similar pixels (a super-pixel). The process of explaining model predictions through LIME starts with the deliberate selection of an instance, or a group of instances (in our case, its super-pixels), that are worthy of interpretation. The selected instance's features are then subtly altered to introduce perturbations, resulting in a diverse set of samples. Subsequently, the black-box model generates predictions for these altered samples.

Since perturbing the entire interpretable representation in our case would require more time and computational resources, we instead perturb a super pixel because certain regions of the images can identify a class; altering whether or not that region contributed to that class can produce the same results as altering the input pixels. Hence, another variant of LIME, viz. Submodular Picks Local interpretable Model-Agnostic Explanations (SP-LIME), which is customized for images. SP-LIME is a methodology designed to tackle the above challenges by employing submodular optimization~\cite{ribeiro2016should}. This approach systematically identifies a collection of representative instances along with their corresponding explanations. Hence, the selection process is essential for improving the clarity and dependability of the model, particularly when there are constraints on computational resources. There are numerous algorithms based on clustering, watershed, energy optimization, and graph methods available for submodular optimisation, specifically for super-pixel segmentation~\cite{kumar2023extensive}. In this study, we have selected the quickshift~\cite{vedaldi2008quick} and SLIC~\cite{achanta2012slic} algorithms to perform the task.

\vspace{-0.2cm}
\section{Experimental Setup}
\label{experimental setup}
In this section, we briefly describe the implementation details and the evaluation metrics used to evaluate the performance of the classification models.

\subsection{\textbf{Implementation Details}}
\label{implementation details}
In this study, we select prominent benchmark models, namely VGG16, VGG19 ~\cite{simonyan2014very}, ResNet50 ~\cite{he2016deep}, Inception V3 ~\cite{szegedy2016rethinking}, DenseNet121, DenseNet201 ~\cite{huang2017densely}, MobileNetV2 ~\cite{sandler2018mobilenetv2}, Xception ~\cite{chollet2017xception}, InceptionResNetV2~\cite{szegedy2017inception}, NASNetLarge, and NASNetMobile~\cite{zoph2018learning} to carry out transfer learning. To conduct a comparative analysis and evaluate their respective performances, both small models with low computational resource requirements and large models with high computational resource requirements are chosen.

Each pre-trained model is enhanced by adding a flattening layer to transform the multi-dimensional output into a one-dimensional vector,a dense layer comprising 1024 neurons with a relu activation function, a dropout rate of 0.25, one more dense layer comprising 512 neurons with a relu activation function, a batch normalisation layer, and again a 0.25 dropout to capture complex relationships within the data. Then, finally, a dense output layer with six neurons (tailored for our task) and a softmax activation function is added to enable multi-class classification, refer to Fig.~\ref{fig:Transfer Learning for Image Classification with LIME} for a visual representation of the transfer learning process for image classification. The customised models are compiled using the adam optimizer with a low learning rate of 0.0001, ensuring fine-grained adjustments during training with categorical cross-entropy as the loss function to optimize the models for accurate multi-class classification. The number of epochs is not predetermined due to each model having its own parameters, hence early stopping is implemented with a patience of five.

The images in our custom dataset are in PNG and JPEG formats and exhibit a variety of dimensions. Hence, process the images by converting them into a NumPy array and reshaping them to a standard dimension of 224 by 224 pixels. Additionally, grayscale images are transformed into RGB format. To facilitate further analysis, we normalise the reshaped images, scaling pixel values from 0 to 1. For the purpose of classification, label encoding is employed, utilising one-hot encoding to represent the categorical labels effectively. The implementation is carried out on a machine equipped with an NVIDIA DGX A100 GPU, but we utilized only 25GB of RAM for training with Google's TensorFlow framework.

\subsection{\textbf{Evaluation Protocols}}
\label{evaluation protocols}

The performance of the developed models is assessed using well-established evaluation metrics, specifically precision, recall, F1-score, and accuracy, as described in~\cite{vujovic2021classification}, ~\cite{grandini2020metrics}. A brief description of each metric, along with its mathematical representation, is provided below:

\noindent\textbf{Accuracy:} It measures the proportion of correctly classified instances among the total instances. Accuracy is straightforward to interpret but can be misleading when classes are imbalanced.

\vspace{-0.25cm}
\begin{equation}
Accuracy =\frac{TP + TN}{TP + FP + FN + TN}
\end{equation}

\noindent\textbf{Precision:} Precision focuses on the relevance of the positive predictions. It is the ratio of true positive predictions to the total number of positive predictions (true positives plus false positives).

\vspace{-0.25cm}
\begin{equation}
Precision =\frac{TP}{TP + FP}
\end{equation}

\noindent\textbf{Recall:} Recall, also known as sensitivity or true positive rate, measures the ability of a classifier to find all the relevant instances. Recall is crucial when missing a positive instance has severe consequences.

\vspace{-0.25cm}
\begin{equation}
Recall = \frac{TP}{TP + FN}
\end{equation}

\noindent\textbf{F1-Score:} The F1-Score is the harmonic mean of precision and recall. It provides a balance between precision and recall, taking both false positives and false negatives into account. F1-Score is particularly useful when classes are imbalanced, as it considers both false positives and false negatives equally.

\vspace{-0.25cm}
\begin{equation}
F1-score = \frac{2 * Recall * Precision}{Recall + Precision}
\end{equation}

These metrics collectively provide insights into the performance of a classification model and are commonly used to evaluate the class-wise effectiveness of the model.

\section{Experimental Results}
\label{experimental results}

In this section, the experimental results of our study are presented. A detailed classification analysis of the different CNN benchmark models using different sampling strategies (random vs. stratified) and data balancing schemes are provided in Section~\ref{classification model random sampling} and Section~\ref{classification model stratified sampling}. Further, the explainer predictions of XAI models LIME and SP-LIME are detailed, along with various ablation studies and state-of-the-art comparisons in Sections ~\ref{explaining the predictions using SP-LIME},~\ref{ablation study}, and~\ref{state of art comparison}, respectively. 

\subsection{\textbf{Classification Model-Random Sampling}}
\label{classification model random sampling}

As mentioned in Section~\ref{data splitting} , we divide the dataset into three segments—train, validation, and test using random sampling approach in which images from the dataset are chosen at random. The forthcoming section explains the outcomes of the unbalanced data vs balanced data-based classification.

\vspace{-0.5cm}
\subsubsection{\textbf{Unbalanced Data}}
\label{unbalanced classification model random sampling}
\begin{figure}[h!]
    \centering
    \footnotesize
    \includegraphics[width =0.45\textwidth]{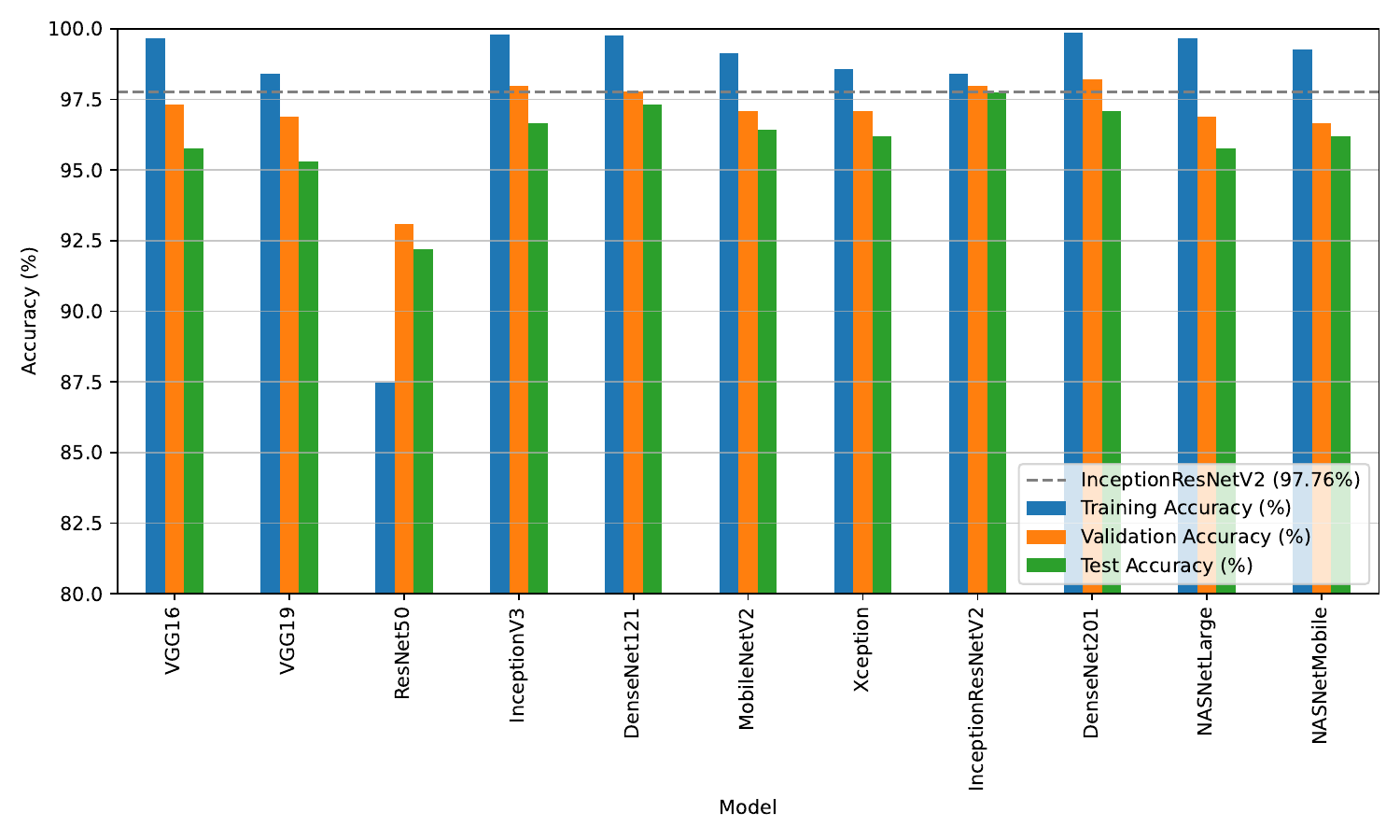}
    \begin{center}
    \captionsetup{font=footnotesize}
    \vspace{-0.30cm}
    \caption{Accuracy of random sampling-based deep learning models using unbalanced data.}
    \label{fig:plt:random_unbalanced_data}
    \end{center}
    \vspace{-0.85cm}
\end{figure}

The random sampling approach on an unbalanced dataset leads to an inconsistent number of images per class in all the splits, which is also non-linear (Refer to Table~\ref{tab:unbalanced random label counts}). The customised deep learning models, initialised with pre-trained weights, undergo rigorous training and evaluation to assess their effectiveness in our classification task. Fig.~\ref{fig:plt:random_unbalanced_data} and Table~\ref{tab:random unbalanced model accuracy} summarises the accuracy achieved with its epoch count by each model during the training process. In particular,  VGG16 -95.75\%, VGG19 - 95.31\%, ResNet50 - 92.18\%, Inception V3 - 96.65\%, DenseNet121 - 97.32\%, MobileNetV2 - 96.42\%, Xception - 96.20\%, InceptionResNetV2 exhibited a high accuracy of 97.76\%, DenseNet201 - 97.09\%, NASNetLarge - 95.75\% and NASNetMobile of 96.20\% are obtained on the test dataset.

\begin{table}[h!]
    \begin{center}
    \renewcommand{\arraystretch}{1.2}
    \setlength\tabcolsep{4pt}
    \footnotesize
    \captionsetup{font=footnotesize}
    \caption{Image counts in the train, validation, and test sets utilizing random sampling approach with unbalanced data.}
    \vspace{0.2cm}
    \begin{tabular}{|c|c|c|c|}
        \hline
        \textbf{Class} & \textbf{Train Set} & \textbf{Validation Set} & \textbf{Test Set} \\
        \hline
        Fish & 959 & 191 & 184 \\
        Human & 27 & 4 & 3 \\
        Mine & 113 & 22 & 32 \\
        Plane & 79 & 27 & 17 \\
        Seafloor & 400 & 95 & 83 \\
        Ship & 514 & 109 & 130 \\
        \hline
        \textbf{Total} & 2092 & 448 & 449 \\
        \hline
    \end{tabular}
    \label{tab:unbalanced random label counts}
    \vspace{-0.65cm}
    \end{center}
\end{table}

\begin{table}[h!]
    \begin{center}
    \footnotesize
    \captionsetup{font=footnotesize}
    \caption{Classification Report for Unbalanced Model using random sampling approach-InceptionResNetV2}
    \vspace{0.2cm}
    \begin{tabular}{|l|cccc|}
    \hline
    \textbf{Class} & \textbf{Precision} & \textbf{Recall} & \textbf{F1-Score} & \textbf{Accuracy} \\
    \hline
    Fish & 1.00 & 1.00 & 1.00 & 1.00 \\
    Human & 1.00 & 1.00 & 1.00 & 1.00 \\
    Mine & 1.00 & 1.00 & 1.00 & 1.00 \\
    Plane & 0.91 & 0.59 & 0.71 & 0.90 \\
    Seafloor & 1.00 & 1.00 & 1.00 & 1.00 \\
    Ship & 0.95 & 0.99 & 0.97 & 0.94 \\
    \hline
    \end{tabular}
    \label{tab:unbalanced classification report random sampling}
    \vspace{-0.65cm}
    \end{center}
\end{table}

\begin{figure}[h!]
    \centering
    \begin{subfigure}{0.225\textwidth}
        \centering
        \includegraphics[width=\linewidth]{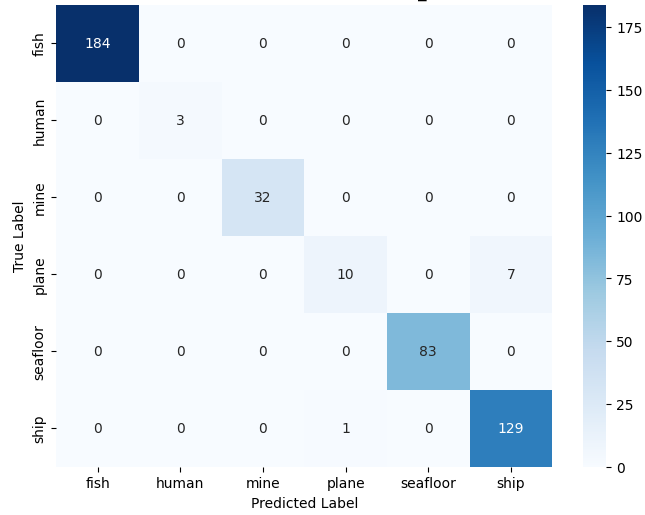}
        \captionsetup{font=footnotesize}
        \caption{InceptionResNetV2 - Unbalanced Model}
        \label{fig:unbalanced_confusion_matrix}
    \end{subfigure}
    \hfill 
    \begin{subfigure}{0.225\textwidth}
        \centering
        \includegraphics[width=\linewidth]{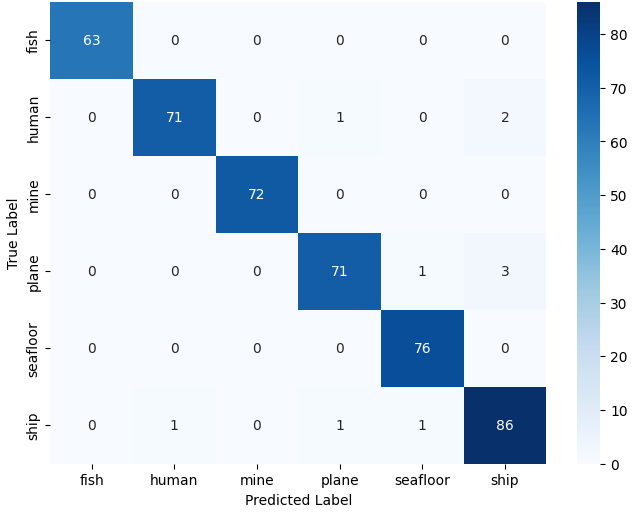}
        \captionsetup{font=footnotesize}
        \subcaption{DenseNet201 - Balanced Model}
        \label{fig:balanced_confusion_matrix}
    \end{subfigure}
    \captionsetup{font=footnotesize}
    \caption{Comparison of confusion matrices between balanced and unbalanced Model using random sampling approach.}
    \label{fig:confusion_matrices_random_sampling}
    \vspace{-0.5cm}
\end{figure}

\begin{table*}[]
    \vspace{-0.75cm}
    \begin{center}
    \renewcommand{\arraystretch}{1.2}
    \footnotesize
    \vspace{0.25cm}
    \captionsetup{font=footnotesize}
    \caption{Accuracy of Deep Learning Models developed utilizing random sampling approaches with unbalanced data.}
    \setlength\tabcolsep{4pt}
    \begin{tabular}{|c|c|c|c|c|}
        \hline
        \textbf{Model} & \textbf{Epochs} & \textbf{Training Accuracy (\%)} & \textbf{Validation Accuracy (\%)} & \textbf{Test Accuracy (\%)} \\
        \hline
        VGG16 & 16 & 99.66 & 97.32 & 95.75 \\
        VGG19 & 6 & 98.42 & 96.88 & 95.31 \\
        ResNet50 & 22 & 87.47 & 93.09 & 92.18 \\
        Inception V3 & 8 & 99.80 & 97.99 & 96.65 \\
        DenseNet121 & 6 & 99.76 & 97.77 & 97.32 \\
        MobileNetV2 & 7 & 99.13 & 97.10 & 96.42 \\
        Xception & 2 & 98.56 & 97.10 & 96.20 \\
        \textbf{InceptionResNetV2} & \textbf{3} & \textbf{98.42} & \textbf{97.99} & \textbf{97.76} \\
        DenseNet201 & 8 & 99.85 & 98.21 &  97.09 \\
        NASNetLarge & 7 & 99.66 & 96.88 & 95.75 \\
        NASNetMobile & 5 & 99.28 & 96.65 & 96.20 \\
        \hline
    \end{tabular}
    \label{tab:random unbalanced model accuracy}
    \vspace{-0.5cm}
    \end{center}
\end{table*}

Although the accuracies are in an acceptable range, from the classification report (Refer to Table~\ref{tab:unbalanced classification report random sampling} and Fig.~\ref{fig:unbalanced_confusion_matrix} it is observed that the test data is suboptimal for some classes such as plane, mine, and human. compared to majority classes of fish, shipwreck, and seafloor images. Such model testing with imbalanced data results in a biased model towards the majority class. To overcome this, data balancing technique i.e. using an equal number of images from each class to train the model to possibly balance the model, as mentioned in ~\cite{kotsiantis2006handling} is envisaged. 

\subsubsection{\textbf{Balanced Data}}
\label{balanced classification model random sampling}

\begin{figure}[h!]
    \centering
    \footnotesize
    \includegraphics[width =0.45\textwidth]{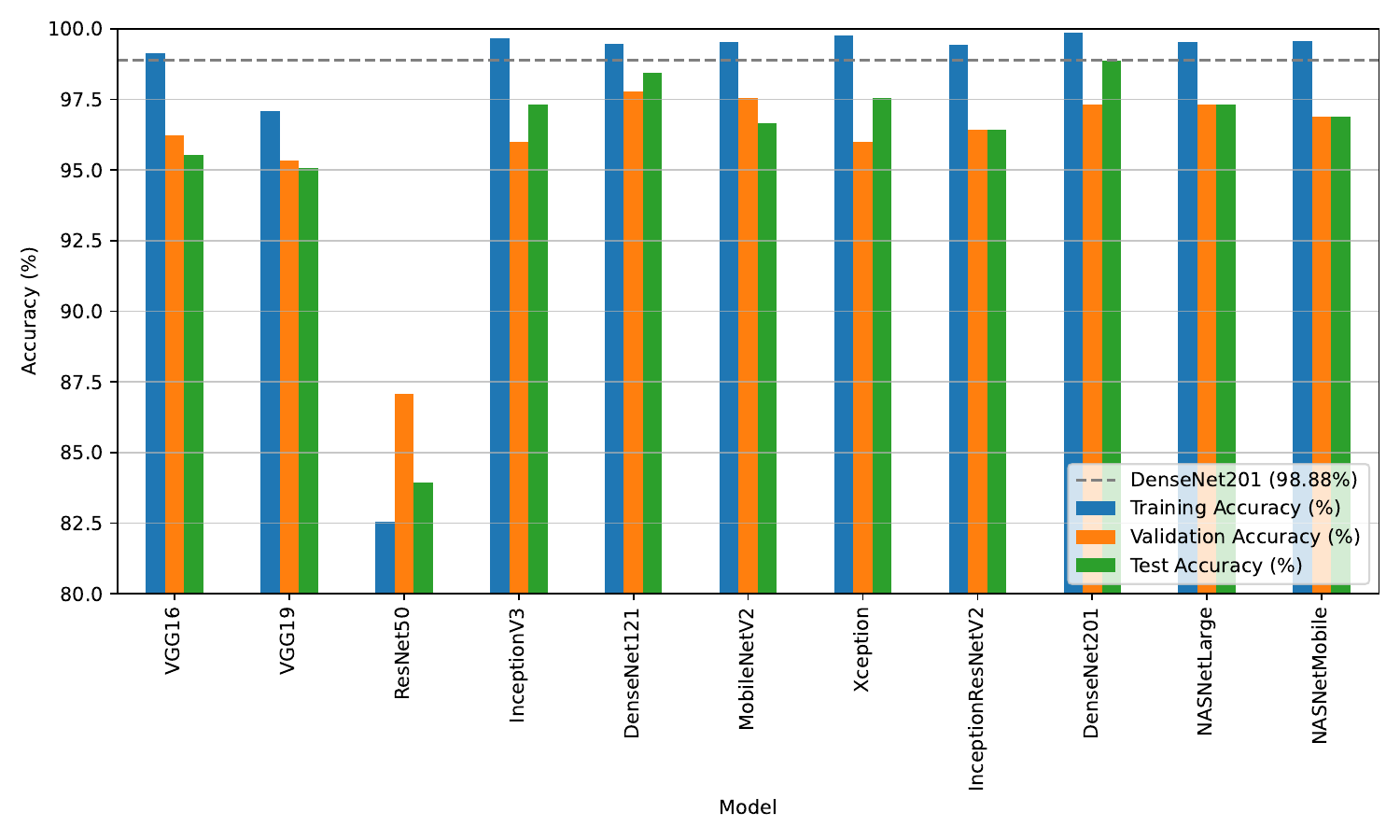}
    \begin{center}
    \captionsetup{font=footnotesize}
    \vspace{-0.35cm}
    \caption{Accuracy of random sampling-based deep learning models using balanced data.}
    \label{fig:plt:random_balanced_data}
    \end{center}
    \vspace{-1.00cm}
\end{figure}

The predictive ability of the models developed in the previous case scenario (unbalanced data) is undoubtedly questionable due to the limited number of test images in the minority class and the abundance of test images in the majority class. To address this issue and achieve data balance across different classes, we employ oversampling and undersampling techniques, as extensively discussed in Section~\ref{handling data imbalance}. As a result, an equal number of 498 images per class is achieved. Further, random sampling approach is utilized to split the dataset into train, validate, and test sets. Hence the number of images per class will be balanced across the splits, as depicted in Table~\ref{tab:random balanced model image count}. The process of image classification using transfer learning is performed again using this newly balanced data. The accuracy achieved by each model during the training process with balanced data along with its epoch count is summarised in Table~\ref{tab: random balanced model accuracy} and in Fig.~\ref{fig:plt:random_balanced_data}. These results highlight the performance of DenseNet201, which achieved an accuracy of 98.88\%. The other backbone models achieved VGG16 - 95.53\%, VGG19 - 95.08\%, ResNet50 - 83.92\%, Inception V3 - 97.32\%, MobileNetV2 - 96.65\%, Xception - 97.54\%, InceptionResNetV2 - 96.42\%, DenseNet121 - 98.43\%, NASNetLarge - 97.32\% and NASNetMobile of 96.87\% accuracies, respectively on the test dataset. 

\begin{table}[h]
    \begin{center}
    \renewcommand{\arraystretch}{1.2}
    \footnotesize
    \captionsetup{font=footnotesize}
    \caption{Image counts in the train, validation, and test sets utilizing random sampling approach with balanced data.}
    \vspace{0.2cm}
    \setlength\tabcolsep{4pt}
    \begin{tabular}{|c|c|c|c|}
        \hline
        \textbf{Class} & \textbf{Train Set} & \textbf{Validation Set} & \textbf{Test Set} \\
        \hline
        Fish & 355 & 80 & 63 \\
        Human & 341 & 83 & 74 \\
        Mine & 357 & 69 & 72 \\
        Plane & 346 & 77 & 75 \\
        Seafloor & 355 & 67 & 76 \\
        Ship & 337 & 72 & 89 \\
        \hline
        \textbf{Total} & 2091 & 448 & 449 \\
        \hline
    \end{tabular} 
    \label{tab:random balanced model image count}
    \vspace{-0.55cm}
    \end{center}
\end{table}

\begin{table}[h!]
    \begin{center}
    \footnotesize
    \captionsetup{font=footnotesize}
    \caption{Classification Report for Balanced Model using random sampling approach-DenseNet201.}
    \vspace{0.2cm}
    \begin{tabular}{|l|cccc|}
    \hline
    \textbf{Class} & \textbf{Precision} & \textbf{Recall} & \textbf{F1-Score} & \textbf{Accuracy} \\
    \hline
    Fish & 1.00 & 1.00 & 1.00 & 1.00 \\
    Human & 0.97 & 0.96 & 0.97 & 0.98 \\
    Mine & 1.00 & 1.00 & 1.00 & 1.00 \\
    Plane & 0.97 & 0.95 & 0.96 & 0.97 \\
    Seafloor & 0.97 & 1.00 & 0.99 & 0.97 \\
    Ship & 0.95 & 0.97 & 0.96 & 0.94 \\
    \hline
    \end{tabular}
    \label{tab:model_accuracy_balanced_data_random_classification_report}
    \vspace{-0.65cm}
    \end{center}
\end{table}

\begin{table*}[h!]
    \vspace{-0.75cm}
    \begin{center}
    \renewcommand{\arraystretch}{1.2}
    \footnotesize
    \captionsetup{font=footnotesize}
    \caption{Accuracy of Deep Learning Models developed utilizing random sampling approaches with balanced data.}
    \setlength\tabcolsep{4pt}
    \begin{tabular}{|c|c|c|c|c|}
        \hline
        \textbf{Model} & \textbf{Epochs} & \textbf{Train Accuracy (\%)} & \textbf{Validation Accuracy (\%)} & \textbf{Test Accuracy (\%)} \\
        \hline
        VGG16 & 14 & 99.13 & 96.21 & 95.53 \\
        VGG19 & 9 & 97.08 & 95.32 & 95.08 \\
        ResNet50 & 51 & 82.54 & 87.08 & 83.92 \\
        InceptionV3 & 7 & 99.66 & 95.99 & 97.32 \\
        DenseNet121 & 8 & 99.47 & 97.77 & 98.43 \\
        MobileNetV2 & 3 & 99.52 & 97.55 & 96.65 \\
        Xception & 17 & 99.76 & 95.99 & 97.54 \\
        InceptionResNetV2 & 3 & 99.43 & 96.43 & 96.42 \\
        \textbf{DenseNet201} & \textbf{12} & \textbf{99.85} & \textbf{97.32} & \textbf{98.88} \\
        NASNetLarge & 4 & 99.52 & 97.32 & 97.32 \\
        NASNetMobile & 6 & 99.56 & 96.88 & 96.87 \\
        \hline
    \end{tabular}
    \label{tab: random balanced model accuracy}
    \vspace{-0.75cm}
    \end{center}
\end{table*}

Although there is a small drop in accuracy after balancing, the performance across all classes is consistent and well-balanced. Specifically, referring to Table~\ref{tab:unbalanced classification report random sampling}, the accuracy of the unbalanced model for the class labels `plane'  is  90\%,  Whereas, with the balanced model- DenseNet201 classification report, the accuracy has improved to 97\%. However, some other class accuracies dropped slightly, e.g. human and seafloor (accuracies dropped from 100\%  to 98\%). However, the model is generalized across all the classes (Refer
Table~\ref{tab:model_accuracy_balanced_data_random_classification_report} and Fig.~\ref{fig:balanced_confusion_matrix}). This demonstrates a substantial improvement of balanced models on individual class performance compared to the performance of unbalanced models.

\subsection{\textbf{Classification Model-Stratified Sampling}}
\label{classification model stratified sampling}
The accuracy of the models developed in the previous Section~\ref{classification model random sampling} leveraged random sampling techniques. As a result, the data sample representation from each class varies in both training and evaluation. To address this issue, an alternative sampling strategy stratified sampling techniques is employed~\cite{parsons2014stratified}. As mentioned in Section~\ref{data splitting}, we divide the dataset into three segments— train, validation, and test using stratified sampling techniques, which ensure that images from all classes are selected with equal probability. The outcomes of the unbalanced data vs balanced data are discussed below:

\subsubsection{\textbf{Unbalanced Data}}
\label{unbalanced classification model stratified sampling}
The stratified sampling approach on an unbalanced dataset leads to a consistent number of images per class in all the splits, which is also linear (Refer to Table~\ref{tab:unbalanced stratified label counts}). 

\begin{table}[h!]
    \begin{center}
    \renewcommand{\arraystretch}{1.2}
    \setlength\tabcolsep{4pt}
    \footnotesize
    \captionsetup{font=footnotesize}
    \caption{Image counts in the train, validation, and test sets utilizing stratified sampling approach with unbalanced data.}
    \vspace{0.2cm}
    \begin{tabular}{|c|c|c|c|}
        \hline
        \textbf{Class} & \textbf{Train Set} & \textbf{Validation Set} & \textbf{Test Set} \\
        \hline
        Fish & 934 & 200 & 200 \\
        Human & 24 & 5 & 5 \\
        Mine & 117 & 25 & 25 \\
        Plane & 86 & 19 & 18 \\
        Seafloor & 404 & 87 & 87 \\
        Ship & 527 & 113 & 113 \\
        \hline
        \textbf{Total} & 2092 & 448 & 449 \\
        \hline
    \end{tabular}
    \label{tab:unbalanced stratified label counts}
    \vspace{-0.85cm}
    \end{center}
\end{table}

\begin{table}[h!]
    \begin{center}
    \footnotesize
    \captionsetup{font=footnotesize}
    \caption{Classification Report for unbalanced Model using stratified sampling approach-DenseNet121.}
    \vspace{0.2cm}
    \begin{tabular}{|l|cccc|}
    \hline
    \textbf{Class} & \textbf{Precision} & \textbf{Recall} & \textbf{F1-Score} & \textbf{Accuracy} \\
    \hline
    Fish & 1.00 & 1.00 & 1.00 & 1.00 \\
    Human & 1.00 & 0.60 & 0.75 & 1.00 \\
    Mine & 1.00 & 1.00 & 1.00 & 1.00 \\
    Plane & 0.93 & 0.78 & 0.85 & 0.93 \\
    Seafloor & 1.00 & 0.99 & 0.99 & 1.00 \\
    Ship & 0.94 & 0.99 & 0.97 & 0.94 \\
    \hline
    \end{tabular}
    \label{tab:classification report stratified unbalanced model}
    \vspace{-0.5cm}
    \end{center}
\end{table}

\begin{figure}[h!]
    \centering
    \footnotesize
    \includegraphics[width =0.49\textwidth]{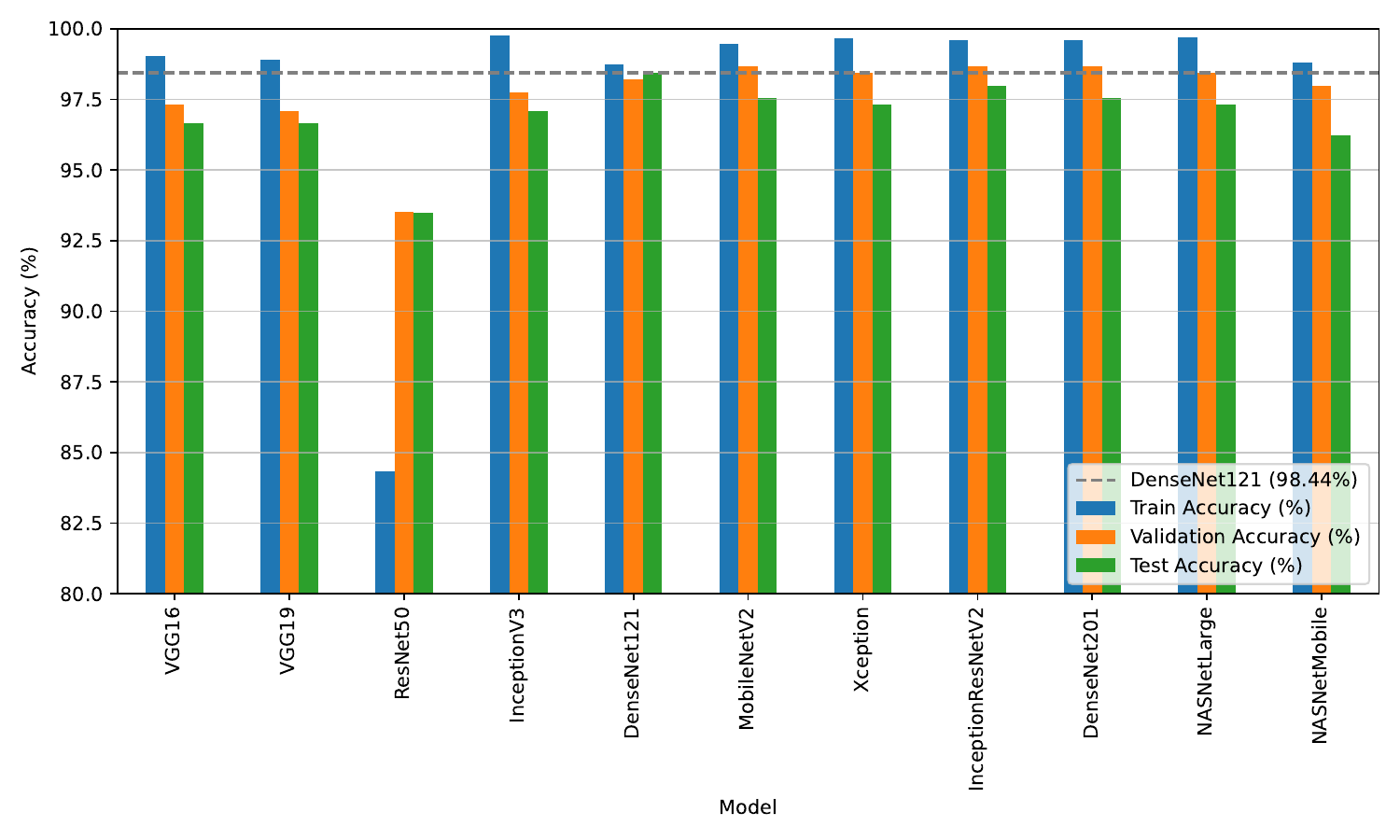}
    \begin{center}
    \captionsetup{font=footnotesize}
    \vspace{-0.5cm}
    \caption{Accuracy of stratified sampling-based deep learning models using unbalanced data.}
    \label{fig:plt:stratified_unbalanced_data}
    \end{center}
    \vspace{-0.85cm}
\end{figure}

The customised deep learning models, initialised with pre-trained weights, undergo rigorous training and evaluation to assess their effectiveness in our classification task. As depicted in the Fig.~\ref{fig:plt:stratified_unbalanced_data}, the following Table~\ref{tab:unbalanced stratified model accuracy} summarises the accuracy achieved by each model, VGG16 achieved 96.65\%, VGG19 - 96.65\%, ResNet50 - 93.50\%, Inception V3 - 97.10\%, DenseNet121 - 98.44\%, MobileNetV2, and DenseNet201 exhibited a accuracy of 97.55\%, Xception - 97.32 \%, InceptionResNetV2 - 97.99,  NASNetMobile -97.32, and NASNetLarge - 96.21\%.

\begin{table*}[]
\vspace{-0.5CM}
    \begin{center}
    \renewcommand{\arraystretch}{1.2}
    \setlength\tabcolsep{4pt}
    \footnotesize
    \captionsetup{font=footnotesize}
    \caption{ Accuracy of Deep Learning Models developed utilizing stratified sampling approaches with unbalanced data.}
    \begin{tabular}{|c|c|c|c|c|}
        \hline
        \textbf{Model} & \textbf{Epochs}  & \textbf{Train Accuracy (\%)} & \textbf{Validation Accuracy (\%)} & \textbf{Test Accuracy (\%)} \\
        \hline
        VGG16 & 9 & 99.04 & 97.32 & 96.65 \\
        VGG19 & 6 & 98.90 & 97.09 & 96.65 \\
        ResNet50 & 18 & 84.32 & 93.52 & 93.50 \\
        Inception V3 & 12 & 99.76 & 97.76 & 97.10 \\
        \textbf{DenseNet121} & \textbf{2} & \textbf{98.75} & \textbf{98.21} & \textbf{98.44} \\
        MobileNetV2 & 4 & 99.47 & 98.66 & 97.55\\
        Xception & 13 & 99.66 & 98.43 & 97.32 \\
        InceptionResNetV2 & 10 & 99.61 & 98.66 & 97.99 \\
        DenseNet201 & 8 & 99.61 & 98.66 & 97.55 \\
        NASNetLarge & 6 & 99.71 & 98.43 & 97.32 \\
        NASNetMobile & 3 & 98.80 & 97.99 & 96.21 \\
        \hline
    \end{tabular}
    \label{tab:unbalanced stratified model accuracy}
    \vspace{-0.75cm}
    \end{center}
\end{table*}

\begin{figure}[h!]
    \centering
    \begin{subfigure}{0.225\textwidth}
        \centering
        \includegraphics[width=\linewidth, trim=0cm 0cm 0cm 0cm, clip]{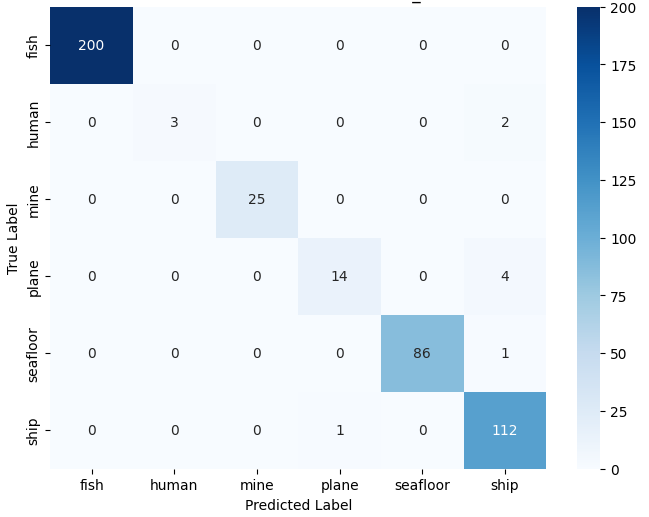}
        \captionsetup{font=footnotesize}
        \caption{DenseNet121 - Unbalanced Model}
        \label{fig:Stratified_unbalanced_confusion_matrix}
    \end{subfigure}
    \hfill 
    \begin{subfigure}{0.225\textwidth}
        \centering
        \includegraphics[width=\linewidth]{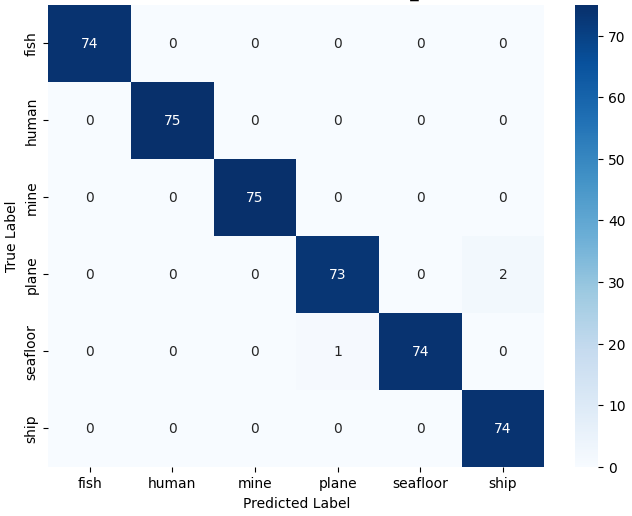}
        \captionsetup{font=footnotesize}
        \caption{DenseNet121 - Balanced Model}
        \label{fig:Stratified_balanced_confusion_matrix}
    \end{subfigure}
    \captionsetup{font=footnotesize}
    \caption{Comparison of confusion matrices between balanced and unbalanced model using stratified sampling approach.}
    \label{fig:confusion_matrices_stratified_sampling}
    \vspace{-0.25cm}
\end{figure}

Although the accuracy is improved compared to the models developed using random sampling approach, the performance of individual classes are still lower compared to the balanced models developed using random sampling techniques (Refer Fig.~\ref{fig:Stratified_unbalanced_confusion_matrix}). By comparing the class-wise accuracies in Table~\ref{tab:unbalanced classification report random sampling} and Table~\ref{tab:classification report stratified unbalanced model}, we can observe a significant increase in individual class performance. The unbalanced model from the random sampling approach achieved an accuracy of 90\% for the class "plane", whereas the counterpart result of the same unbalanced data through a stratified sampling approach achieves an accuracy of 93\%, which is considerably higher but not equivalent to the accuracy of other classes.

\subsubsection{\textbf{Balanced Data}}
\label{balanced classification model stratified sampling}

The stratified approach improved the overall accuracy, but the individual class accuracy is still lower. Hence, the predictive ability of the models developed above utilizing stratified sampling approach is still undoubtedly questionable due to the limited number of test images in the minority class and the abundance of test images in the majority class. To address this issue and to achieve data balance across different classes, oversampling and undersampling techniques are carried out, as extensively discussed in Section~\ref{handling data imbalance}. As a result, 498 images per class is achieved, and stratified sampling approach is utilized in spitting the dataset into train, validate, and test sets. The number of images per class after splitting get balanced, as mentioned in Table~\ref{tab:balanced stratified label counts}.

\begin{table}[h!]
    \vspace{-0.2cm}
    \begin{center}
    \renewcommand{\arraystretch}{1.2}
    \footnotesize
    \captionsetup{font=footnotesize}
    \caption{Image counts in the train, validation, and test sets utilizing stratified sampling approach with balanced data.}
    \vspace{0.2cm}
    \setlength\tabcolsep{4pt}
    \begin{tabular}{|c|c|c|c|}
        \hline
        \textbf{Class} & \textbf{Train Set} & \textbf{Validation Set} & \textbf{Test Set} \\
        \hline
        Fish & 349 & 75 & 74 \\
        Human & 348 & 75 & 75 \\
        Mine & 348 & 75 & 75 \\
        Plane & 348 & 75 & 75 \\
        Seafloor & 349 & 74 & 75 \\
        Ship & 349 & 75 & 74 \\
        \hline
        \textbf{Total} & 2091 & 448 & 449 \\
        \hline
    \end{tabular}
    \label{tab:balanced stratified label counts}
    \vspace{-0.5cm}
    \end{center}
\end{table}

\begin{figure}[h!]
    \centering
    \footnotesize
    \includegraphics[width =0.45\textwidth]{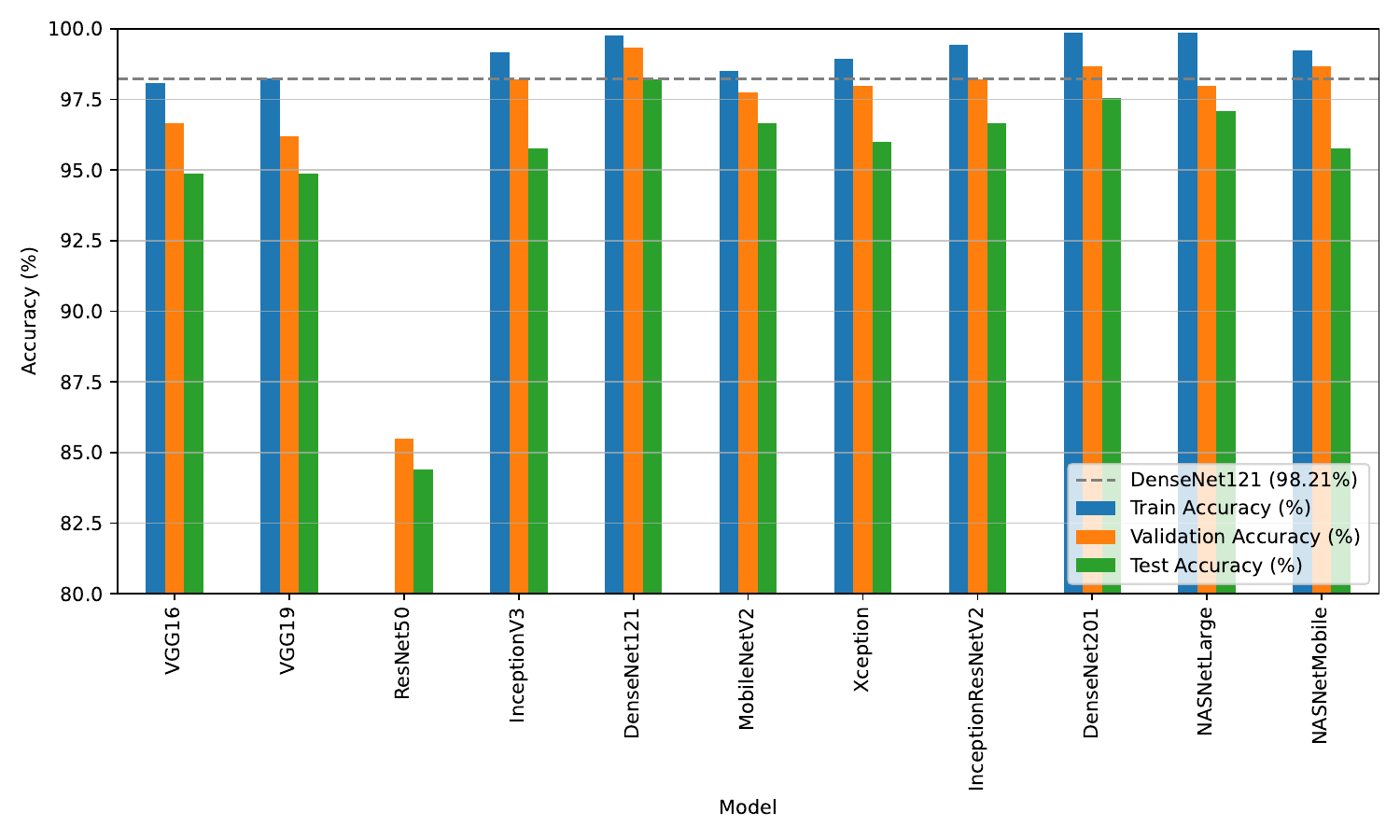}
    \begin{center}
    \captionsetup{font=footnotesize}
    \vspace{-0.5cm}
    \caption{Accuracy of stratified sampling-based deep learning models using balanced data.}
    \label{fig:plt:stratified_balanced_data}
    \end{center}
    \vspace{-0.85cm}
\end{figure}

The process of image classification using transfer learning is performed using this balanced data using stratified sampling. The accuracy achieved by each model during the training process with balanced data on the test dataset and validation dataset is summarised in Table~\ref{tab:stratified balanceded model accuracy} and depicted in the Fig.~\ref{fig:plt:stratified_balanced_data}. These results highlight the performance of DenseNet121, which achieved an impressive accuracy of 98.21\%, DenseNet201 - 97.55\%, NASNetLarge - 97.10\%, MobileNetV2 and InceptionResNetV2 - 96.65\%, Xception - 95.76\%, InceptionV3 and NASNetMobile - 95.66\%. VGG19 - 94.87\%, VGG16 - 94.87 and ResNet50 - 84.40\%, all surpassing the 95\% accuracy mark. 

\begin{table}[h]
    \begin{center}
    \footnotesize
    \captionsetup{font=footnotesize}
    \caption{Classification Report for balanced Model using stratified sampling approach-DenseNet121.}
    \vspace{0.2cm}
    \begin{tabular}{|l|cccc|}
    \hline
    \textbf{Class} & \textbf{Precision} & \textbf{Recall} & \textbf{F1-Score} & \textbf{Accuracy} \\
    \hline
    Fish & 1.00 & 1.00 & 1.00 & 1.00 \\
    Human & 1.00 & 1.00 & 1.00 & 1.00 \\
    Mine & 1.00 & 1.00 & 1.00 & 1.00 \\
    Plane & 0.99 & 0.97 & 0.98 & 0.98 \\
    Seafloor & 1.00 & 0.99 & 0.99 & 1.00 \\
    Ship & 0.97 & 1.00 & 0.99 & 0.97 \\
    \hline
    \end{tabular}
    \label{tab:classification report balanced model stratified approach}
    \vspace{-0.5cm}
    \end{center}
\end{table}

\begin{table*}[]
    \begin{center}
    \renewcommand{\arraystretch}{1.2}
    \footnotesize
    \captionsetup{font=footnotesize}
    \caption{Accuracy of Deep Learning Models developed utilizing stratified sampling approaches with balanced data.}
    \setlength\tabcolsep{4pt}
    \begin{tabular}{|c|c|c|c|c|}
        \hline
        \textbf{Model} & \textbf{Epochs} & \textbf{Train Accuracy (\%)} & \textbf{Validation Accuracy (\%)} & \textbf{Test Accuracy (\%)} \\
        \hline
        VGG16 & 8 & 98.08 &  96.65 & 94.87 \\
        VGG19 & 12 & 98.23 & 96.20 & 94.87 \\
        ResNet50 & 37 & 72.40 & 85.49 & 84.40 \\
        Inception V3 & 5 & 99.18 & 98.21 & 95.76 \\
        \textbf{DenseNet121} & \textbf{15}& \textbf{99.76} & \textbf{99.33} & \textbf{98.21} \\
        MobileNetV2 & 2 & 98.51 & 97.76 & 96.65 \\
        Xception & 3 & 98.94 & 97.99 & 95.99 \\
        InceptionResNetV2 & 12 & 99.42 & 98.21 & 96.65 \\
        DenseNet201 & 7 & 99.85 & 98.66 & 97.55 \\
        NASNetLarge & 5 & 99.85 & 97.99 & 97.10 \\
        NASNetMobile & 6 & 99.23 & 98.66 & 95.76 \\
        \hline
    \end{tabular}
    \label{tab:stratified balanceded model accuracy}
    \vspace{-0.75cm}
    \end{center}
\end{table*}

The classification report, as shown in Table~\ref{tab:classification report balanced model stratified approach} and the Fig.~\ref{fig:Stratified_balanced_confusion_matrix}, indicates that the class-wise accuracy has improved. By using a stratified approach on balanced data, the accuracy for ``human", ``plane" and ``ship" has reached 100\%, 98\% and 97\% respectively. This represents an improvement from the previous model's accuracy developed using balanced data through random sampling techniques for ``plane" 
(from 93\% to 98\%) and  for ``ship" (from 94\% to 97\%). These insights emphasize the critical role of choosing the correct sampling approach and balanced data along with well-suited, pre-trained models. Among the options considered, \textbf{DenseNet121 from the stratified approach using balanced data} stood out as the most fitting choice for our classification task, demonstrating outstanding validation accuracy of 99.33\% and a test accuracy of 98.21\%, hence it is considered the de-facto model for our following XAI study.

\subsection{\textbf{Explaining the predictions from the classifier using SP-LIME}}
\label{explaining the predictions using SP-LIME}
 
As a result of various combinations of random vs. stratified sampling strategies as well as balanced and unbalanced approaches (as mentioned in previous Sections~\ref{classification model random sampling} and~\ref{classification model stratified sampling}), we developed altogether forty-four distinct classification models, with forty of them having an accuracy of more than 95\%. We chose the model DenseNet121 with the highest accuracy and superior generalisation abilities for each class developed using stratified approach on balanced data and incorporated the post-hoc SP-LIME technique, a variation of LIME, to enhance explainability. 

Although LIME can provide valuable insights, it is important to carefully choose which instances perturb, since perturbing all the pixels in the image demands more time and computation, hence selected instances are perturbed such as super-pixels, \textit{viz.} SP-LIME. This is because users may not have the time or capacity to review a large number of explanations, as explained in Section~\ref{section: lime as a xai tool}.

\begin{figure}[h!]
    \centering
    \includegraphics[width=0.45\textwidth]{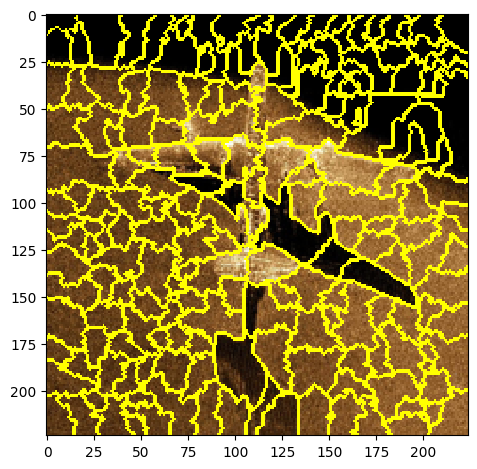}
    \vspace{-0.45cm}
    \begin{center}
        \captionsetup{font=footnotesize}
        \caption{Super-pixels from QuickShift Algorithm}
        \label{fig:super-pixels}
    \end{center}
    \vspace{-0.5cm}
\end{figure}

\begin{figure*}[h!]
    \vspace{-0.5cm}
    \centering
    \includegraphics[width=\textwidth]{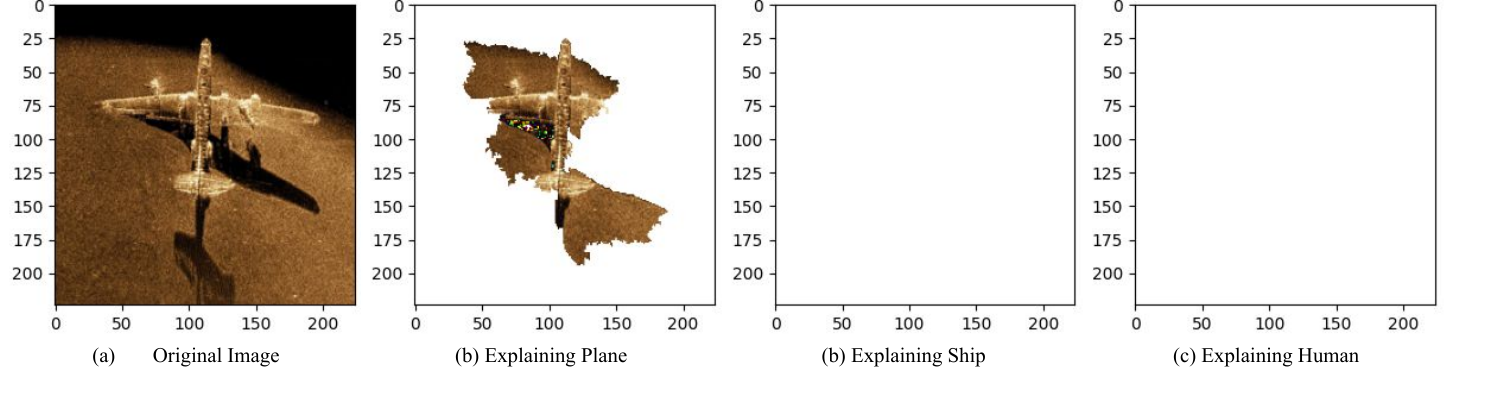}
    \begin{minipage}{\textwidth}
        \captionsetup{font=footnotesize}
        \caption{Explaining the prediction made by DenseNet121 neural network (Balanced Model) using SP-LIME. The top 3 classes predicted are "Plane" (p = 0.100), "Ship" (p = 0.000) and "Human" (p = 0.000)}
        \label{balanced model with LIME}
    \end{minipage}
    \vspace{-0.75cm}
\end{figure*}

The hyper-parameters of the explainer is the number of features and the samples. Features refer to the characteristics or attributes of data points such as pixel values, colour channels, and texture information, whereas samples are individual data points or instances in the dataset. Ten features are taken into account, and 300 samples are used for the analysis with QuickShift~\cite{vedaldi2008quick} algorithm for submodular optimization, it considers features (such as pixel intensities) to identify regions of similar density. The quickShift algorithm with a kernel size of 2, a maximum distance of 100, and a ratio of 0.1 then groups samples (pixels) based on their proximity and density, produces 186 super-pixels, as illustrated in the Fig.~\ref{fig:super-pixels}. From the visual explanation in Fig.~\ref{balanced model with LIME}, for the top three predicted classes and their associated probabilities: `Plane' with a probability of 1.00000, `Ship' with 0.00000, and `Human' with 0.00000, it is clear that the plane's features, including the tail, head, and parts of the wings, are carefully considered for image classification. Conversely, the shadow region is disregarded, indicating that these areas do not contribute to the classification. The probabilities for all other classes are zero, hence the explainer does not mask any region and it points out that the explainer operation is solely focused on the classification model. This comparative analysis illuminates the nuanced performance and interpretability of SP-LIME explainer. 

The explainer process for the above model took 16.99 seconds for sampling, contributing to a total execution time of 149.88 seconds. To accommodate varying computational resource demands, we have the flexibility to reduce the number of super-pixels. The determination of the optimal number of features and perturbations is contingent upon the inherent complexity of the images under consideration. When appropriately tuned, it has the capacity to yield exceptional results within a reduced timeframe and with minimal computational overhead. This adaptability underscores the efficiency of SP-LIME in generating accurate and interpretable explanations for complex image data. The presented visual representations provide valuable insights into the explainer's functioning, which is intricately tied to the underlying model predictions which underscores the significance of leveraging SP-LIME explainer as a tool that empowers users to make informed decisions based on a clearer understanding of model predictions through visual representations.

\subsection{\textbf{Ablation Study}}
\label{ablation study}
In this section, various ablation studies such as comparing the performance of the explainer with a balanced and an unbalanced model, evaluating the impact of different hyper-parameters and assessing the computation time and performance for LIME and SP-LIME across different classes are carried out.

\subsubsection{\textbf{Explainer with balanced and unbalanced models}}
\label{explainer with balanced and unbalanced models}
In this experiment, the impact of the SP-LIME explainer with sampling and data balancing strategies is analysed. Both the random and stratified sampling along with balanced vs unbalanced datasets using our top-performing model are studied. Table~\ref{tab:SP-LIME with different models} summarises the visual explanation along with the original image.

\begin{table*}[h!]
    \begin{center}
    \footnotesize
    \captionsetup{font=footnotesize}
    \caption{SP-LIME with different models: First row: sampling techniques, second row: model's balance, third-row predicted class: plane, and fourth-row predicted class: ship, and the corresponding visual explanation images from that particular class.}
    \vspace{0.2cm}
    \begin{tabular}{l}
    \includegraphics[width=0.95\textwidth]{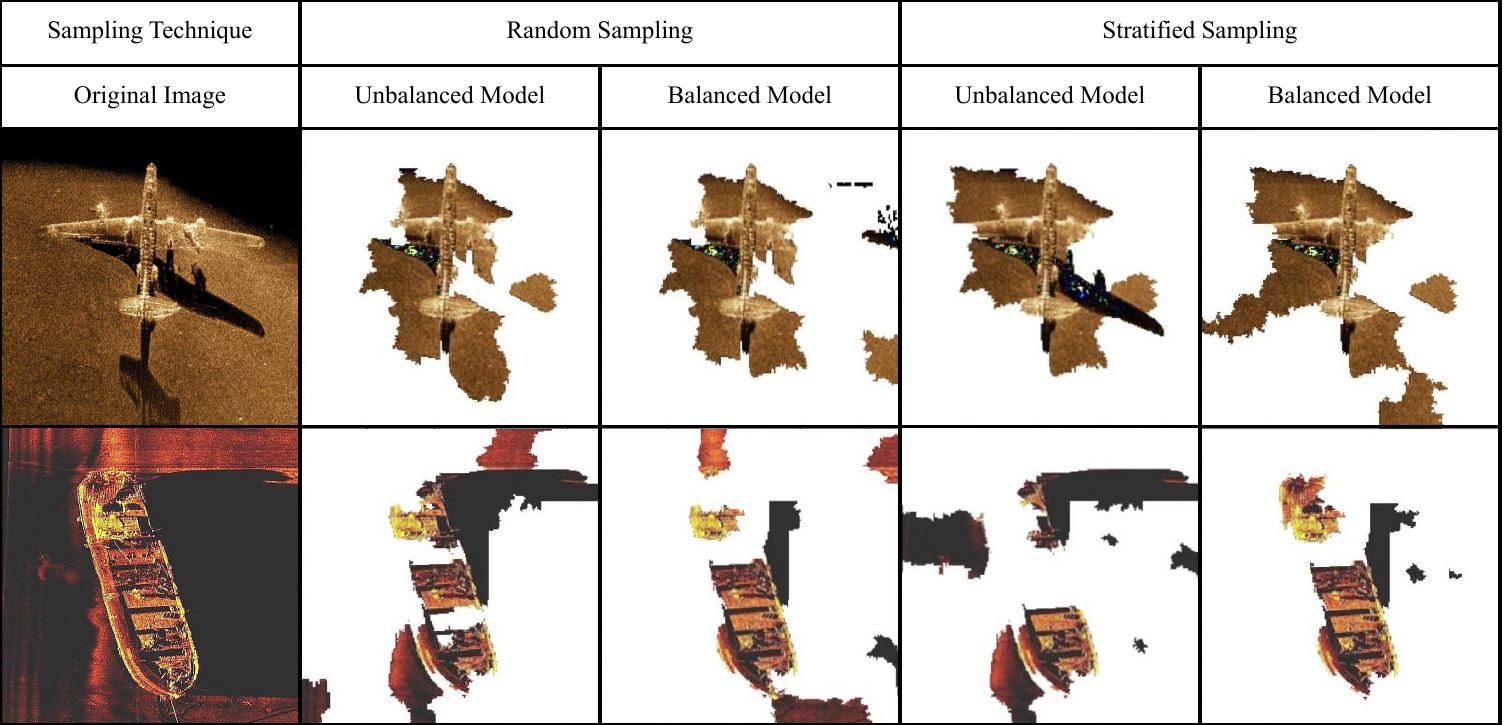} \\
    \end{tabular}
    \label{tab:SP-LIME with different models}
    \end{center}
    \vspace{-0.4cm}
\end{table*}

Table~\ref{tab:SP-LIME with different models}, it can be observed that the \textit{balanced model from the stratified sampling approach} provides the best visual explanation, with a focus on the area with the exact mask. Compared to the random sampling technique, stratified sampling outperformed in selecting the target object precisely. Similarly, it is seen that the unbalanced model from both approaches results in more shadow regions, compared to the balanced model. For the plane and ship prediction, it is clear that the mask from the stratified balanced model outperforms the other models in explaining the tail\&wing and the relevant ship regions, respectively.

\subsubsection{\textbf{Hyper-parameter of the Explainer: Number of Features}}
\label{hyper-parameters of the explainer -  features}

This section tries to understand the impact of number of features as a hyperparameter towards LIME explanation. Since SP-LIME perturbs at the super-pixel level while LIME perturbs each pixel, we assess the computational time taken for each approach, to better understand the performances and computational requirements of LIME and SP-LIME. 

\begin{table}[h!]
    \vspace{-0.15cm}
    \centering
    \footnotesize
    \captionsetup{font=footnotesize}
    \caption{Comparison with 100 Samples on LIME \& SP-LIME}
    \vspace{0.2cm}
    \begin{tabular}{|c|c|c|c|}
    \hline
    \multirow{2}{*}{\textbf{Number of Features}} & \multicolumn{2}{c|}{\textbf{Computation Time in Seconds}} \\
    \cline{2-3}
    & \textbf{LIME} & \textbf{SP-LIME} \\ 
    \hline
         10 & 5.98 & 5.20 \\  
         25 & 6.75 & 6.09 \\
         50 & 6.61 & 5.87 \\
         100 & 6.38 & 7.03 \\
         400 & 6.87 & 7.21 \\
         500 & 5.87 & 7.09 \\
    \hline
    \end{tabular}
    \label{tab:comparison with 100 samples}
    \vspace{-0.5cm}
\end{table}

\begin{figure}[h!]%
        \centering
        \begin{subfigure}[htbp]{0.25\textwidth}
            \centering
            \fbox{\includegraphics[width=3.2cm]{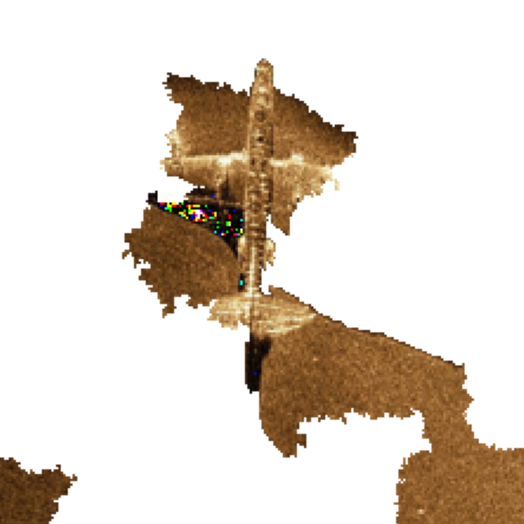}}%
            \caption{}
        \end{subfigure}%
        \begin{subfigure}[htbp]{0.25\textwidth}
            \centering
            \fbox{\includegraphics[width=3.2cm]{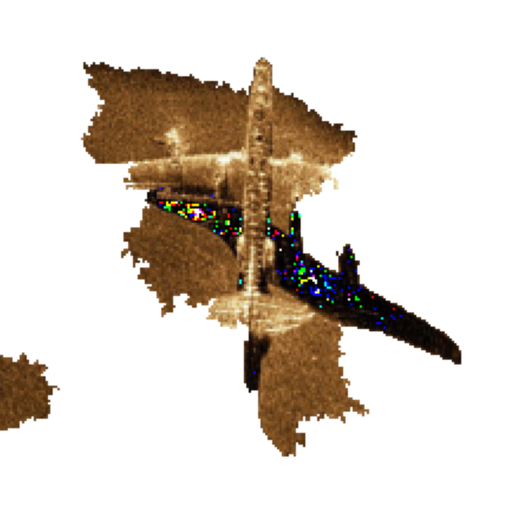}}%
            \caption{}
        \end{subfigure}\break\vskip 1mm
        \begin{subfigure}[htbp]{0.25\textwidth}
            \centering
        \fbox{\includegraphics[width=3.2cm]{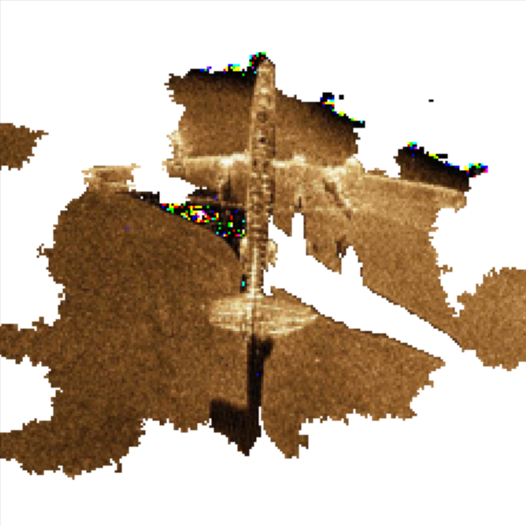}}%
            \caption{}
        \end{subfigure}%
        \begin{subfigure}[htbp]{0.25\textwidth}
            \centering
        \fbox{\includegraphics[width=3.2cm]{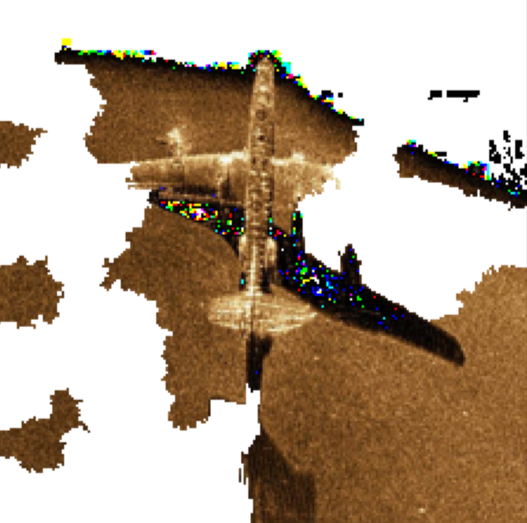}}%
            \caption{}
        \end{subfigure}%
        \captionsetup{font=footnotesize}
        \caption{Comparison of LIME and SP-LIME explanations on varing number of features: (a) 10 features \& 100 samples on LIME, (b) 10 features \& 100 samples on SP-LIME, (c) 500 features \& 100 samples on LIME, (d) 500 features \& 100 samples on SP-LIME.}
        \label{n features vs 100 samples}
        \vspace{-0.4cm}
\end{figure}

We test LIME and SP-LIME's performance in a number of different parameter settings, specifically investigating the number of features, tweaking to 10, 25, 50, 100, 150, 200, 250, 300, 400, and 500. Although SP-LIME has its own hyper-parameters, we have considered the following parameters as fixed as follows: super-pixel algorithm: quickshift, kernel size: 2, max distance: 100, and ratio: 0.1. The number of features denote the quantity of features or input variables that the explainer will perturb in order to produce a local explanation. The absence of a large class group, coupled with relatively low variability in the dataset, diminishes the computational expense sensitivity to changes in the number of features for both LIME and SP-LIME. Despite this, the choice of the feature count significantly influences the effectiveness of the explanations provided for the classification decisions (refer to Fig.~\ref{n features vs 100 samples}) and Table~\ref{tab:comparison with 100 samples}.

From Table~\ref{tab:comparison with 100 samples}, it can be observed that if more features are considered for both LIME and SP-LIME, then it may result in more computational time requirement spent calculating the mask. The comparative visualization analysis of LIME and SP-LIME by varying the number of features is given in Fig.~\ref{n features vs 100 samples}. It can be observed that Fig.~\ref{n features vs 100 samples} (a) from LIME and Fig.~\ref{n features vs 100 samples} (b) from SP-LIME with 10 features outperform the rest (c) and (d) with 500 features. Further, SP-LIME takes less time to compute but gives a comparable explanation to LIME, which can be further tuned with some changes in parameters.

\subsubsection{\textbf{Hyper-parameter of the Explainer: Number of Samples}}
\label{hyper-parameters of the explainer - samples}

The number of samples represents the number of synthetic samples generated by the explainer for each perturbation. We fix the number of features to 10, vary the number of samples with values ranging from 100 to 1000, and analyse the performance of our XAI model. The computational time associated with each combination of parameters are shown in Table~\ref{tab:comparison with 10 features}.

\begin{figure}[h!]
    \centering
    \begin{subfigure}[htbp]{0.25\textwidth}
    \centering
        \fbox{\includegraphics[width=3.2cm]{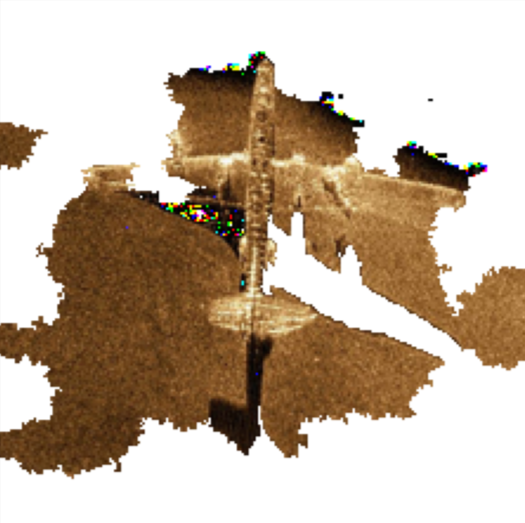}}%
       \caption{}
    \end{subfigure}%
    \begin{subfigure}[htbp]{0.25\textwidth}
    \centering
        \fbox{\includegraphics[width=3.2cm]{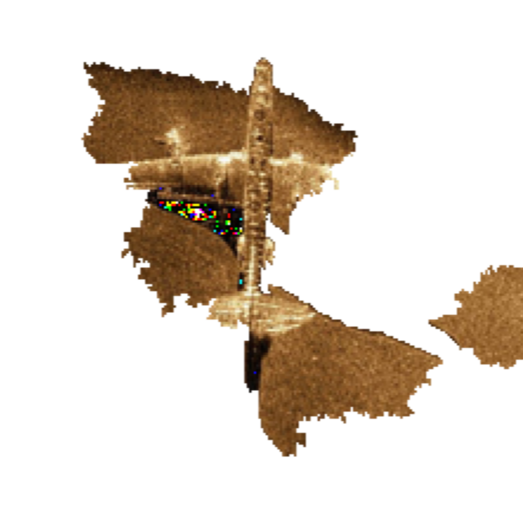}}%
    \caption{}
    \end{subfigure}%
    \captionsetup{font=footnotesize}
    \caption{Comparison of LIME and SP-LIME explanations on varing number of samples: (a) 10 features \& 1000 samples on LIME, (b) 10 features \& 1000 samples on SP-LIME.}
    \vspace{-0.25cm}
    \label{10 features vs n samples}
\end{figure}

\begin{table}[h!]
    \centering
    \footnotesize
    \captionsetup{font=footnotesize}
    \caption{Comparison with 10 Features on LIME \& SP-LIME.}
    \vspace{0.2cm}
    \begin{tabular}{|c|c|c|}
    \hline
    \multirow{2}{*}{\textbf{Number of Samples}} & \multicolumn{2}{c|}{\textbf{Computation Time in Seconds}} \\
    \cline{2-3}
    & \textbf{LIME} & \textbf{SP-LIME} \\ 
    \hline
         100 & 5.98 & 5.20 \\
         200 & 13.42 & 10.40 \\
         300 & 18.84 & 13.95 \\
         800 & 52.55 & 40.42 \\
         900 & 53.03 & 46.78 \\
         1000 & 66.83 & 52.70 \\
    \hline
    \end{tabular}
    \label{tab:comparison with 10 features}
    \vspace{-0.5cm}
\end{table}

\begin{figure}[h!]
    \captionsetup{font=footnotesize}
    \centering
    \begin{subfigure}[htbp]{0.25\textwidth}
        \centering
        \fbox{\includegraphics[width=3.2cm]{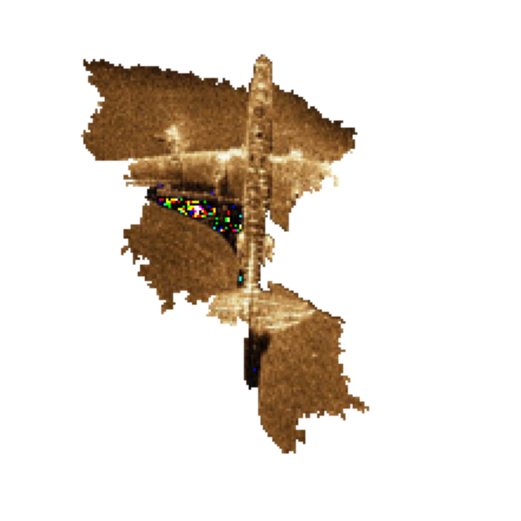}}%
        \caption{}
    \end{subfigure}%
    \begin{subfigure}[htbp]{0.25\textwidth}
    \centering
        \fbox{\includegraphics[width=3.2cm]{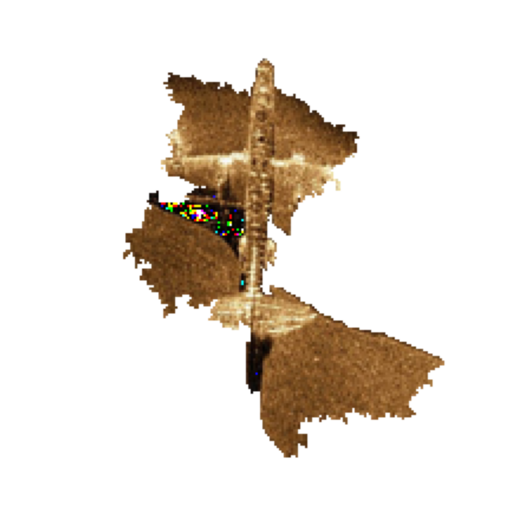}}%
        \caption{}
    \end{subfigure}%
    \caption{A Comparison of Effective Explanations from SP-LIME and LIME: (a) 10 features \& 500 samples on LIME, (b) 10 features \& 500 samples on SP-LIME.}
    \vspace{-0.25cm}
    \label{LIME vs sp-LIME}
\end{figure}

From Table~\ref{tab:comparison with 10 features}, it can be observed that the number of samples also influences the explainer, which results in an increase in computation with an increase in the number of samples. From Fig.~\ref{10 features vs n samples}, it can be noted that the explanation from SP-LIME is better than LIME since the LIME captured more regions, which is of no interest with more computational resources and more time, but on the other hand, the SP-LIME masked the zone, which is very close to the region of interest. Depending on the dataset and the computational constraints available, a trade-off between the number of features and samples should be made for a better explanation. So we chose 10 features and 500 samples and applied them to both LIME and SP-LIME. The LIME explainer took 31.61 seconds. On the other hand, the SP-LIME explainer took 20 seconds and produced comparable results (refer to Fig.~\ref{LIME vs sp-LIME}).

Without sacrificing the interpretability that LIME achieves, SP-LIME offers a faster runtime. The specified parameters for SP-LIME were crucial in achieving this efficiency for our dataset, demonstrating the effectiveness of the chosen configuration. This comparison provides valuable insights into the trade-offs between runtime and interpretability in the context of model explanation techniques.

\subsubsection{\textbf{SP-LIME and the corresponding Super-pixel Algorithm}}
\label{sp-lime with various optimization algorithms}

\begin{figure}[h!]%
    \captionsetup{font=footnotesize}
    \centering
    \begin{subfigure}[htbp]{0.25\textwidth}
        \centering
        \fbox{\includegraphics[width=3.2cm]{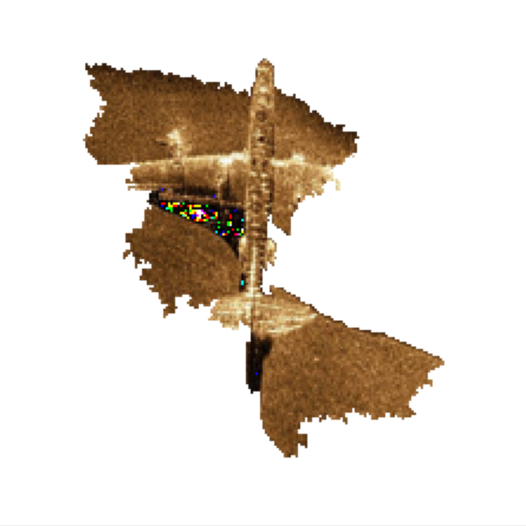}}%
        \caption{}
    \end{subfigure}%
    \begin{subfigure}[htbp]{0.25\textwidth}
        \centering
        \fbox{\includegraphics[width=3.2cm]{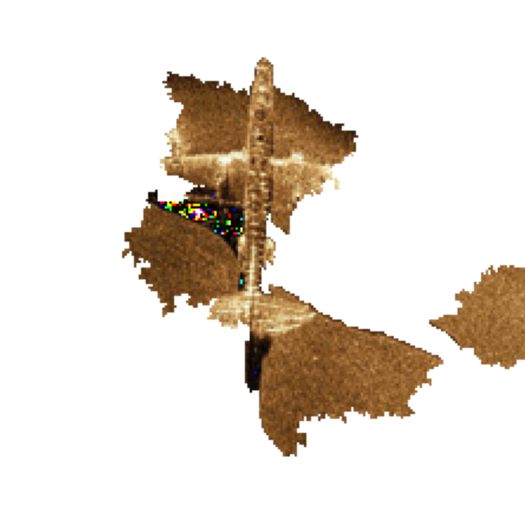}}%
        \caption{}
    \end{subfigure}%
    \caption{Comparing  explanations from distinct super-pixel algorithms on SP-LIME: (a) Quickshift, (b) SLIC (Simple Linear Iterative Clustering.)}
    \vspace{-0.25cm}
    \label{quickshift vs slic}
\end{figure}

The algorithm used to generate super-pixels also affects SP-LIME's performance. So, we conducted a brief analysis on the performance of SP-LIME, with a specific focus on two distinct algorithms: quickshift~\cite{vedaldi2008quick} and SLIC~\cite{achanta2012slic} (Simple Linear Iterative Clustering). For the quickshift algorithm, the SP-LIME explainer took 20.00 seconds, with key parameters including a kernel size of 2, a maximum distance of 100, a ratio of 0.1, 10 selected features, and 500 samples. On the other hand, the SLIC algorithm took 20.47 seconds with similar parameter configurations. Notably, the findings highlight the computational efficiency and performance of SP-LIME (refer figure~\ref{quickshift vs slic}) when employing the quickshift algorithm in contrast to SLIC, shedding light on the nuanced trade-offs between computational resources and the quality of decision explanations. 

From Fig.~\ref{quickshift vs slic}, it can be noticed that both explanations are similar to each other, however, the explanation using quickshift with SP-LIME captured more regions in the target object that SLIC did not capture. The computational time difference between both algorithms is minimal, but for our dataset, Quickshift performs better.

\subsubsection{\textbf{Impact of explainer on different classes}}
\label{explainers impact on different classes}

\begin{figure}[h!]
    \captionsetup{font=footnotesize}
    \centering
    \begin{subfigure}[b]{0.3\textwidth}
        \centering
        \includegraphics[width=\textwidth]{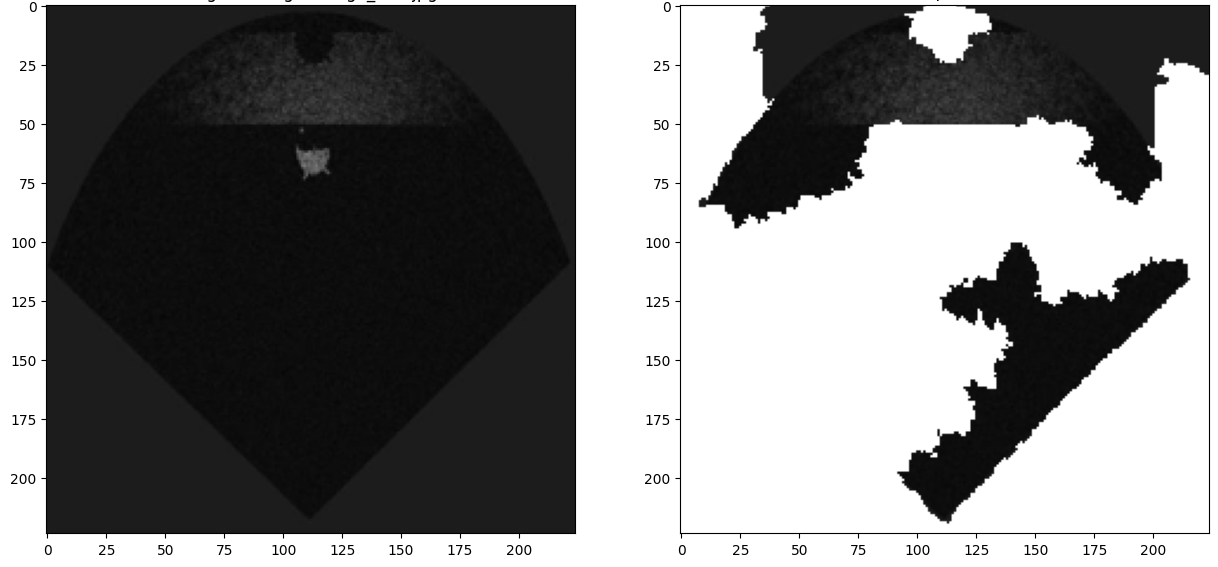}
        \caption{Explanation for Mine}
        \label{fig:Mine explanation}
    \end{subfigure}
    \hfill
    \begin{subfigure}[b]{0.3\textwidth}
        \centering
        \includegraphics[width=\textwidth]{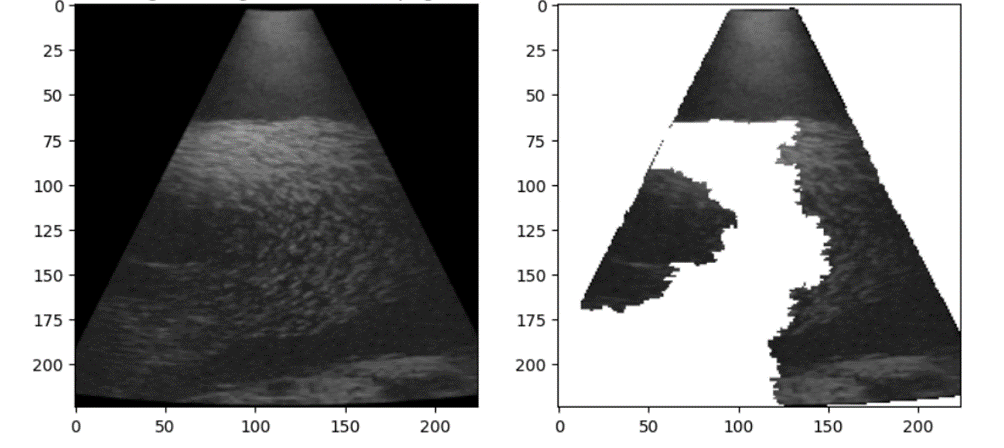}
        \caption{Explanation for Fishes}
        \label{fig:fish explanation}
    \end{subfigure}
    \hfill
    \begin{subfigure}[b]{0.3\textwidth}
        \centering
        \includegraphics[width=\textwidth]{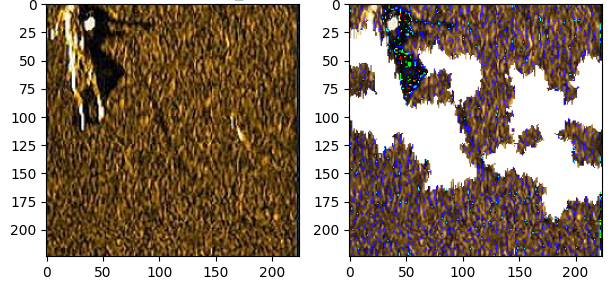}
        \caption{Explanation for Human}
        \label{fig:human explanation}
    \end{subfigure}
    \caption{Inaccurate Explanations}
    \vspace{-0.25cm}
    \label{fig:inaccurate explanations}
\end{figure}

The quality, variety, and veracity of the images used during training have a significant impact on the explainer's effectiveness. Analysis of the classes: mine, fish, and human from the Fig.~\ref{fig:Mine explanation}, Fig.~\ref{fig:fish explanation}, and Fig.~\ref{fig:human explanation} suggests that the classification model's accuracy is compromised, evident in the inaccuracies of the generated masks that fail to align with the actual regions of interest.

\begin{figure}[h!]
    \captionsetup{font=footnotesize}
    \centering
    \begin{subfigure}[b]{0.3\textwidth}
        \centering
        \includegraphics[width=\textwidth]{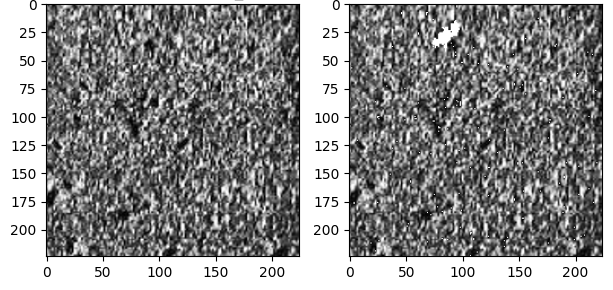}
        \caption{Explanation for Seafloor}
        \label{fig:seafloor explanation}
    \end{subfigure}
    \hfill
    \begin{subfigure}[b]{0.3\textwidth}
        \centering
        \includegraphics[width=\textwidth]{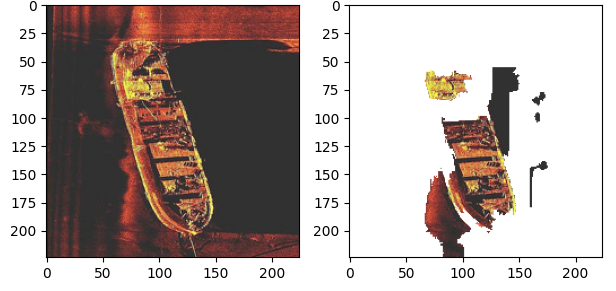}
        \caption{Explanation for Ship}
        \label{fig:ship explanation}
    \end{subfigure}
    \hfill
    \begin{subfigure}[b]{0.3\textwidth}
        \centering
        \includegraphics[width=\textwidth]{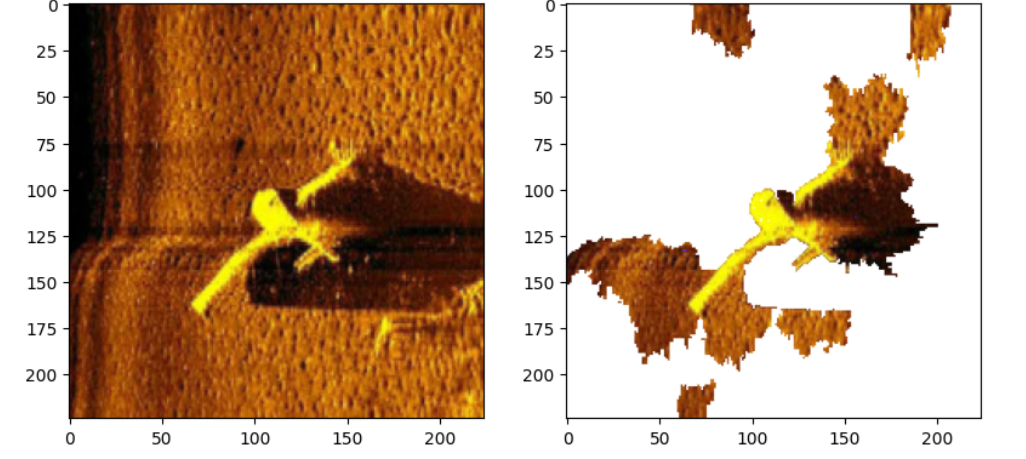}
        \caption{Explanation for Plane}
        \label{fig:plane explanation}
    \end{subfigure}
    \caption{Accurate Explanations}
    \vspace{-0.5cm}
    \label{fig:accurate explanations}
\end{figure}

The explanation mask accuracy differs due to training data. The original ``human" dataset had 34 images, and augmentation added 498 images to this dataset. However, this augmentation may have caused the images to differ, causing the model to learn patterns rather than accurately capture human objects' intrinsic shapes. Hence, the model's performance suffers, and the explainer struggles to provide clear explanations. For the `fish' and `mine' classes, the dataset is of poor quality with data limitations and similarity, so the model failed to provide meaningful explanations. On the other hand, for the classes plane, ship, and seafloor (refer Fig.~\ref{fig:seafloor explanation}, Fig.~\ref{fig:ship explanation}, and Fig.~\ref{fig:plane explanation}), promising explanations are achieved due of the quality and quantity of images used for training. We envisage that image enhancement techniques such as SRGAN~\cite{wang2018esrgan}  and attention techniques could be employed to enhance image quality and to direct explainer towards the area of interest.


\subsection{State of the Art Comparison}
\label{state of art comparison}

To showcase the efficacy of our classification model, a thorough comparison was conducted with existing models, evaluating their respective accuracies. In underwater SONAR imagery, there are very few research works as follows, Divya et al.~\cite{divyabarathi2021object} employed ResNet50 and ensemble techniques on seabed KLSG data, achieving a notable accuracy of 90\%. In a study by Najib et al.~\cite{najibzadeh2023active} on NAT III, a DCNN yielded an impressive accuracy of 98.44\%. Wang et al.~\cite{wang2019underwater}, utilizing AWCNN, achieved an accuracy of 86.90\%. Nguyen et al.~\cite{nguyen2019study}, working with CKI and TDI datasets, utilized GoogLeNet and attained an accuracy of 91.60\%. Chungath et al.~\cite{chungath2023transfer}, combining seabed and custom datasets and employing ResNet34, reported an accuracy of 97.33\%. Fernandes et al.~\cite{fernandes2022deep} utilized a tailor-made CNN on a custom dataset, achieving a maximum accuracy of 99\%. Luo et al.~\cite{luo2019sediment}, using a shallow CNN with seabed sediment data, obtained an accuracy of 94.40\%. Zhou et al.~\cite{zhou2021research} employed SCTD and achieved an accuracy of 95.84 through IMDNet. The results, as illustrated in Table~\ref{tab:state of art comparison image classification}, unequivocally highlight the superior performance of our customised DenseNet121 model, positioning it at the forefront of the evaluated models with an accuracy of 98.21\%.

\begin{table}[t!]
    \footnotesize
    \captionsetup{font=footnotesize}
    \caption{State of the Art (SOTA) Comparison - Image Classifier.}
    \centering
    \setlength{\tabcolsep}{3.7pt} 
    \begin{tabular}{|c|c|c|}
    \hline
        \textbf{Model} & \textbf{Dataset} & \textbf{Accuracy (\%)} \\
        \hline
        ResNet50~\cite{divyabarathi2021object} & Seabed Objects KLSG & 90.00 \\
        DCNN~\cite{najibzadeh2023active} & Array Technology III (NAT III) & 98.44 \\
        AWCNN~\cite{wang2019underwater} & Custom Dataset & 86.90 \\
        GoogleNet~\cite{nguyen2019study} & CKI, TDI-2017 and TDI-2018 & 91.60 \\
        ResNet34~\cite{chungath2023transfer} & Seabed + Custom Dataset & 97.33 \\
        CNN~\cite{fernandes2022deep} & Custom Dataset & 99.00 \\
        Shallow CNN~\cite{luo2019sediment} & Seabed sediment & 94.40 \\
        IMDNet~\cite{zhou2021research} & SCTD & 95.84 \\
        VGG16~\cite{du2023comparative} & Seabed Objects KLSG & 94.81 \\
        ResNet50~\cite{fuchs2018object} & Aracati & 88.00 \\
        \hline
        \textbf{DenseNet121 (Ours)} & \textbf{Custom Dataset} & \textbf{98.21} \\
    \hline
    \end{tabular}
    \label{tab:state of art comparison image classification}
    \vspace{-0.25cm}
\end{table}

\begin{table*}[h!]
    \footnotesize
    \centering
    \captionsetup{font=footnotesize}
    \caption{State of the Art (SOTA) Comparison: Explainer (XAI).}
    \begin{tabular}{|p{10cm}|p{2.5cm}|p{2.5cm}|}
    \hline
    \textbf{Existing Paper} & \textbf{Dataset} & \textbf{Explainer} \\
    \hline
        Explainable Systematic Analysis For Synthetic Aperture SONAR Imagery~\cite{walker2021explainable} &  Multi-site and Single-site & LIME \\
    \hline
        Deep learning-based explainable target classification for synthetic aperture radar images~\cite{pannu2020deep} & MSTAR & LIME\\
    \hline
        LIME-Assisted Automatic Target Recognition with SAR Images: Towards Incremental Learning and Explainability~\cite{oveis2023LIME} & MSTAR & LIME \\
    \hline
        Underwater Sonar Image Classification and Analysis using LIME-based Explainable Artificial Intelligence & Custom Dataset & LIME and SP-LIME \\
    \hline
    \end{tabular}
    \label{tab:state of art comparison explainer}
    \vspace{-0.5cm}
\end{table*}

For the comparison of our explainer in our research paper, we reviewed three key papers that contribute to the field of explainable AI for synthetic aperture imagery, tabulated in Table~\ref{tab:state of art comparison explainer}. In the field of SONAR imagery, Walker et al.~\cite{walker2021explainable} used LIME to learn about important features that play a part in improving performance and to learn how important it is to balance data for fine-tuning deep learning models in classifying SAS seafloor images. In 2023, Oveis et al.~\cite{oveis2023LIME} conducted a study using LIME, for target recognition with synthetic radar images. Additionally, Pannu et al.~\cite{pannu2020deep} also utilized LIME for a similar purpose. Utilizing radar images, Pannu et al.'s (2020) study used LIME to perform target classification, Both studies utilized the MSTAR dataset. Since there are no specific metrics for explaining the model, the usefulness of these explainer models depends entirely on the user's perspective. We utilized LIME to provide explanations for the predictions made by an underwater SONAR image classifier, thereby improving the trust and effectiveness of the black box model.

\section{Conclusion and Future Works}
\label{conclusion}

In this study, our primary objective was to develop effective deep learning methodologies to tackle the inherent classification challenges associated with underwater acoustic images. Due to the limited availability of public datasets in this specialized domain, we curated a customized dataset by amalgamating data from multiple sources. This tailored dataset formation emerged as a crucial facet of our investigation. Our methodology centers around using a pre-trained deep neural network. We examined various backbone models, such as VGG16, VGG19, ResNet50, InceptionV3, DenseNet121, MobileNetV2, Xception, InceptionResNetV2, DenseNet201, NASNetLarge, and NASNetMobile. We applied both stratified and random sampling approaches, as well as on balanced and unbalanced datasets. Through a combination of different sampling strategies and data balancing techniques, we found that the best model was DenseNet121. We fine-tuned this powerful architecture to our specific task using the limited data at our disposal. The experimental results unequivocally demonstrated the efficacy of our transfer learning strategy, showcasing remarkable outcomes in the classification of underwater acoustic images. A noteworthy observation arising from our study was the potency of well-established sampling techniques in addressing challenges associated with imbalanced datasets. Rather than resorting to resource-intensive and intricate synthetic data generation methods, we illustrated the effectiveness of careful dataset balancing through sampling. This strategic approach significantly mitigated misclassifications, especially for classes characterized by a scarcity of samples. To enhance the interpretability of our classification model, we integrated the explainer models LIME and SP-LIME. This addition not only elevates the transparency of the model but also provides users with a clearer understanding of predictions. The incorporation of SP-LIME contributes to instilling confidence in users and facilitates more informed decision-making based on the model's outputs.
Notably, the SP-LIME mask reveals a slight misfit with the object of interest, even though the model is trained on class labels. To address this, our forthcoming work aims to refine the masking accuracy through the integration of image enhancement techniques such as Super-Resolution Generative Adversarial Networks (SRGAN)~\cite{wang2018esrgan}, focusing the Explainer on the region of interest, and the incorporation of richer and more diverse datasets with an expanded range of classes. This promising avenue holds the potential to further augment the model's interpretability and predictive accuracy.

\bibliographystyle{ieee}
\bibliography{SONAR_LIME}

\end{document}